%% file: main.tex
\documentclass[peerreview,11pt]{IEEEtran}
\IEEEoverridecommandlockouts

\usepackage[utf8]{inputenc}
\usepackage[T1]{fontenc}
\usepackage{amsmath,amssymb,amsfonts,amsthm}
\usepackage{algorithm}
\usepackage[noend]{algpseudocode}
\usepackage{enumitem}
\usepackage{caption}
\usepackage{subcaption}
\usepackage{graphicx}
\usepackage{textcomp}
\usepackage{xcolor}
\usepackage[square,numbers]{natbib}

\newtheorem{definition}{Definition}
\newtheorem{remark}{Remark}
\newtheorem{proposition}{Proposition}

\newcommand{\ldq}{\text{``}}
\newcommand{\rdq}{\text{''}}
\newcommand{\algcomment}[1]{{\small\color{gray}~~// {#1}}}

\newcommand{\changedSB}[1] {{#1}} 
\newcommand{\changedAS}[1] {{#1}} 
\newcommand{\changedASnew}[1] {{#1}} 

\makeatletter
\def\BState{\State\hskip-\ALG@thistlm}
\makeatother
\begin{document}

\title{
\changedSB{Topo-Geometrically Distinct Path Computation}
using Neighborhood-augmented Graph, and its Application to Path Planning for a Tethered Robot in 3D}
\author{Alp Sahin\thanks{Department of Mechanical Engineering and Mechanics, Lehigh University, 19 Memorial Drive West, Bethlehem, PA 18015, U.S.A., \texttt{[als421,sub216]@lehigh.edu}.} ~and~ Subhrajit Bhattacharya \vspace{-2.5em}}

\maketitle
\vspace{3em}

\begin{abstract}
Many robotics applications benefit from being able to compute multiple locally optimal paths in a given configuration space. Existing paradigm is to use topological path planning, which can compute optimal paths in distinct topological classes. However, these methods usually require non-trivial geometric constructions which are prohibitively expensive in 3D, and are unable to distinguish between distinct topologically equivalent geodesics that are created due to high-cost/curvature regions or genus-zero obstacles in 3D. In this paper we propose a universal and generalized approach to computing multiple, locally-optimal paths using the concept of a novel neighborhood-augmented graph, on which graph search algorithms can compute multiple optimal paths that are topo-geometrically distinct. Our approach does not require complex geometric constructions, and the resulting paths are not restricted to distinct topological classes, making the algorithm suitable for problems where locally optimal and geometrically distinct paths are of interest. We demonstrate the application of our algorithm to planning shortest traversible paths for a tethered robot in 3D with cable-length constraint.
\end{abstract}


\section{Introduction}

\input{sections/introduction.tex}


\changedSB{

\section{Technical Background, Related Literature and Problem Motivation}

\subsection{Discrete Path Planning Algorithms}


Discrete search based path planning has been used extensively in solving motion planning problems in robotics because of its simplicity, effectiveness and efficiency~\cite{Ferguson_2008_6207,UrbanChallenge:VT:09,Ben:PR2:planning,conf-icra-SwaminathanPL15}.
It involves an incremental and on-the-fly construction of a discrete representation of a configuration space (usually a graph representation that is created systematically or through sampling, or a simplicial complex representation), and exploration of the discrete representation using a search algorithm from a start state until a target or goal state is reached.
Traditional graph search algorithms such as Dijkstra's \cite{dijkstra-dijkstra}, A* \cite{Hart-Astar} and D* \cite{Stentz-dstar} construct and explore a configuration graph in a systematic manner, while algorithms such as PRM~\cite{PRM:Kavraki:96} and RRT~\cite{lavalle:rrt:2001} use sampling based construction and randomized search approaches.
In recent years, development of any-angle search algorithms~\cite{Daniel:JAIR:10,ijcai:CuiHG17} have allowed computation of optimal paths that are not necessarily restricted to a graph, and development of search algorithms for simplicial complex representations (instead of graph representations) have allowed computation of smooth paths that are optimal in the underlying configuration space~\cite{Bhattacharya:SStar:19}.

\subsection{Prior Work: Topological Path Planning -- Homotopy Invariants and Homotopy-augmented Graph}

The main theoretical and algorithmic contribution of this paper is to develop an algorithm for computing multiple distinct paths from a given start to a given goal point in a configuration. To motivate this work, in this section we start by introducing some related prior work on computation of multiple paths in \emph{topologically distinct} classes in a configuration space, and discuss some of its applications.
Fore more details on this topic, the reader can refer to the author's prior work on topological path planning~\cite{ICRA:14:tethered,cable:separation:IJRR:14,wang2022congestion,planning:AURO:12,Homotopy-Planning:journal:18}.

\begin{definition}[Homotopy Classes of Paths]
    Two paths connecting the same start and goal points in a configuration space are said to be in the same \emph{homotopy class} (or \emph{homotopic}) if one can be continuously deformed into another (without intersecting/crossing obstacles (Figure~\ref{fig:homotopy_def}). Otherwise they are called \emph{non-homotopic}.
\end{definition}
Homotopy classes are the main topological classes of interest when it comes to paths or trajectories in a configuration space.
A \emph{homotopy invariant} is a quantitative measure of homotopy classes, which is a function of paths in the configuration space, and is represented as $h(\gamma)$ for a path $\gamma$. The construction of such homotoy invariants require geometric constructions in the configuration space (for example, non-intersecting \emph{rays} emanating from obstacles in planar domains, or surfaces bounded by obstacles in spatial domains -- Figures~\ref{fig:main_figure}(b,c)).
Homotopy invariants can be incorporated in constructing discrete search graphs called \emph{$h$-augmented graphs}, and search-based planning algorithms can be employed for computating (locally) shortest path in different homotopy classes using a process referred to as \emph{topological path planning} (TPP).

Some of the applications of the ability to compute locally shortest paths in distinct topological classes enabled by TPP include path planning for tethered robot with cable length constraints~\cite{ICRA:14:tethered}, object separation using a cable attached to two robots at its ends~\cite{cable:separation:IJRR:14}, workspace \& motion planning for a multiple-cable controlled robot~\cite{Wang:Cable-Controlled:2018}, and multi-agent path planning with topological reasoning where agents distribute in an environment based on the available topological classes~\cite{DARS:14:HRI, Kim:IROS13,wang2022congestion}.

\subsection{Locally-optimal Paths}

In this section we formalize the notion of \emph{locally-optimal paths} (also referred to as geodesic paths) and make some key observations about them.

\begin{definition}[Locally-optimal or Geodesic Paths~\cite{gromov1999metric}]
    A path connecting a fixed pair of start and goal points in a path metric space (a space along with a \emph{cost/length function} for measuring the cost/length of a path) is called \emph{locally optimal} or \emph{geodesic} if any small (infinitesimal) perturbation to the path results in an increase in the cost of the path.
\end{definition}

We first observe that given a start and a goal point in a configuration space, there can be multiple distinct geodesics connecting them that may or may not be in different homotopy classes (Figure~\ref{fig:main_figure}).
The presence of distinct homotopy classes give rise to distinct geodesic paths (at least one in each homotopy class)~\cite{do1992riemannian}. But even within the same homotopy class there can be multiple geodesic paths created due to geometry, curvature and non-uniform cost.

\begin{remark}[\textbf{Key Observation about Tangents to Geodesics}]
Two distinct geodesic paths connecting the same start and goal points in a configuration space have at least one point common to both the paths 
at which the tangents to the paths are distinct or diverge from each other resulting in the paths to separate from (i.e., not overlap) each other.
\end{remark}
Usually the said common point on geodesic paths where the tangents do not match or diverge from each other is the goal point (Figure~\ref{fig:top_geo}(a)), $q_g$.
However, it is possible for distinct geodesic paths to have parallel tangents at the goal, in which case there exists a point, $q$, on the paths earlier than the goal where the tangents are not parallel/diverge (Figure~\ref{fig:top_geo}(b)).
In that case, one can consider $q$ to be an intermediate goal, where the tangents to the paths are distinct or they diverge from each other (and hence there are two distinct geodesic paths to $q$), and then the overlapping geodesic path from $q$ to $q_g$ simply concatenated to those distinct geodesic paths to obtain distinct geodesic paths connecting $q_s$ and $q_g$.


\begin{remark}[\textbf{Path Neighborhoods as Proxy for Tangent Vectors}]
    Since a reasonable representation of a tangent space of a point is a small neighborhood around the point in the configuration space, tangent vector to a path at that point can be approximately represented by a neighborhood around the path close to the point (see Figure~\ref{fig:top_geo}).
    We call such a representative neighborhood a \emph{path neighborhood} of the path at the point.
\end{remark}

\begin{figure}[h!]
    \centering
    \includegraphics[width=0.95\linewidth]{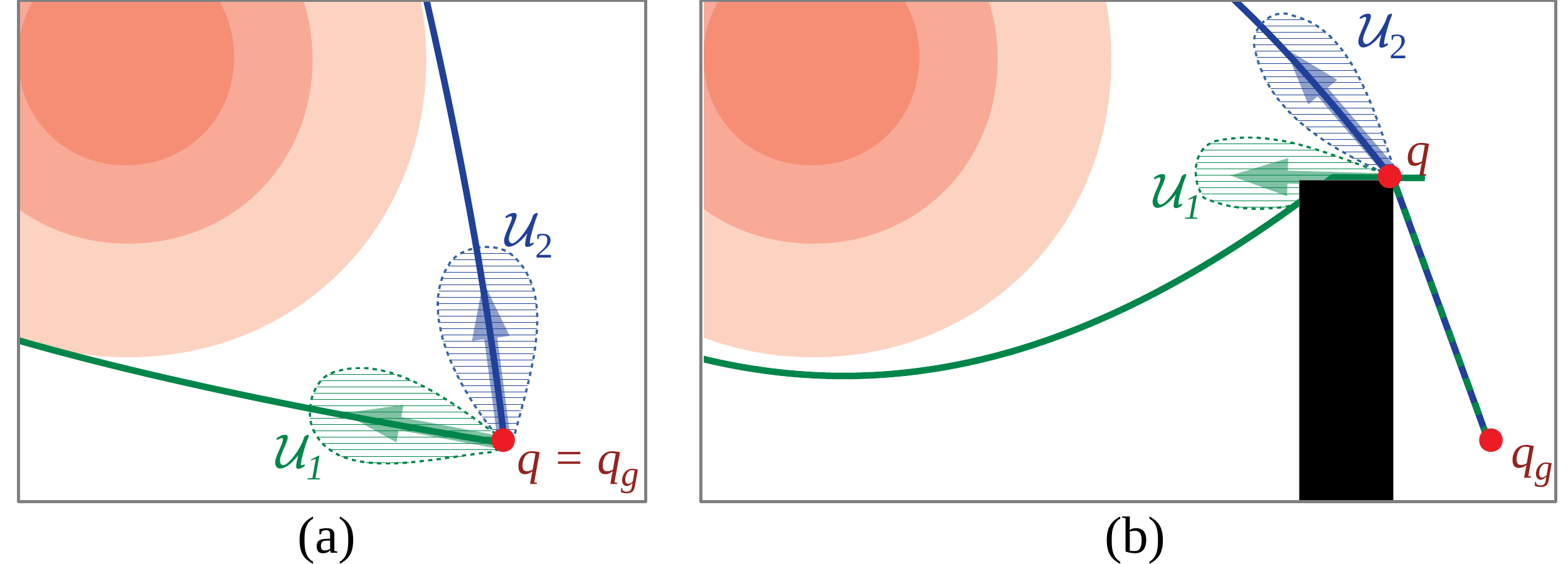}
    \caption{\changedSB{Optimal/geodesic paths to a point $q_g$ in the configuration space that are topo-geometrically distinct have distinct tangent vectors (arrows) at a point $q$ (with $q=q_g$ in case (a), but a point other than $q_g$ in case (b)). Instead of comparing and distinguishing the paths by their tangent vectors (which is often computationally difficult), a reasonable approach is to compare the \emph{path neighborhoods} (shown by hatched/shaded regions) in the proximity of the paths near the point $q$.}}
    \label{fig:top_geo}
\end{figure}

Motivated by these observations we provide the following definition:

\begin{definition}[Topo-geometrically Distinct Paths]\label{def:topo-geometric}
Two distinct geodesic paths connecting the same start and target points, $q_s$ and $q_g$, are called \emph{topo-geometrically distinct} if 
there exists a common point, $q$, on the paths at which 
they have distinct path neighborhoods.
\end{definition}

\begin{figure}
    \centering
    \includegraphics[width=0.95\linewidth]{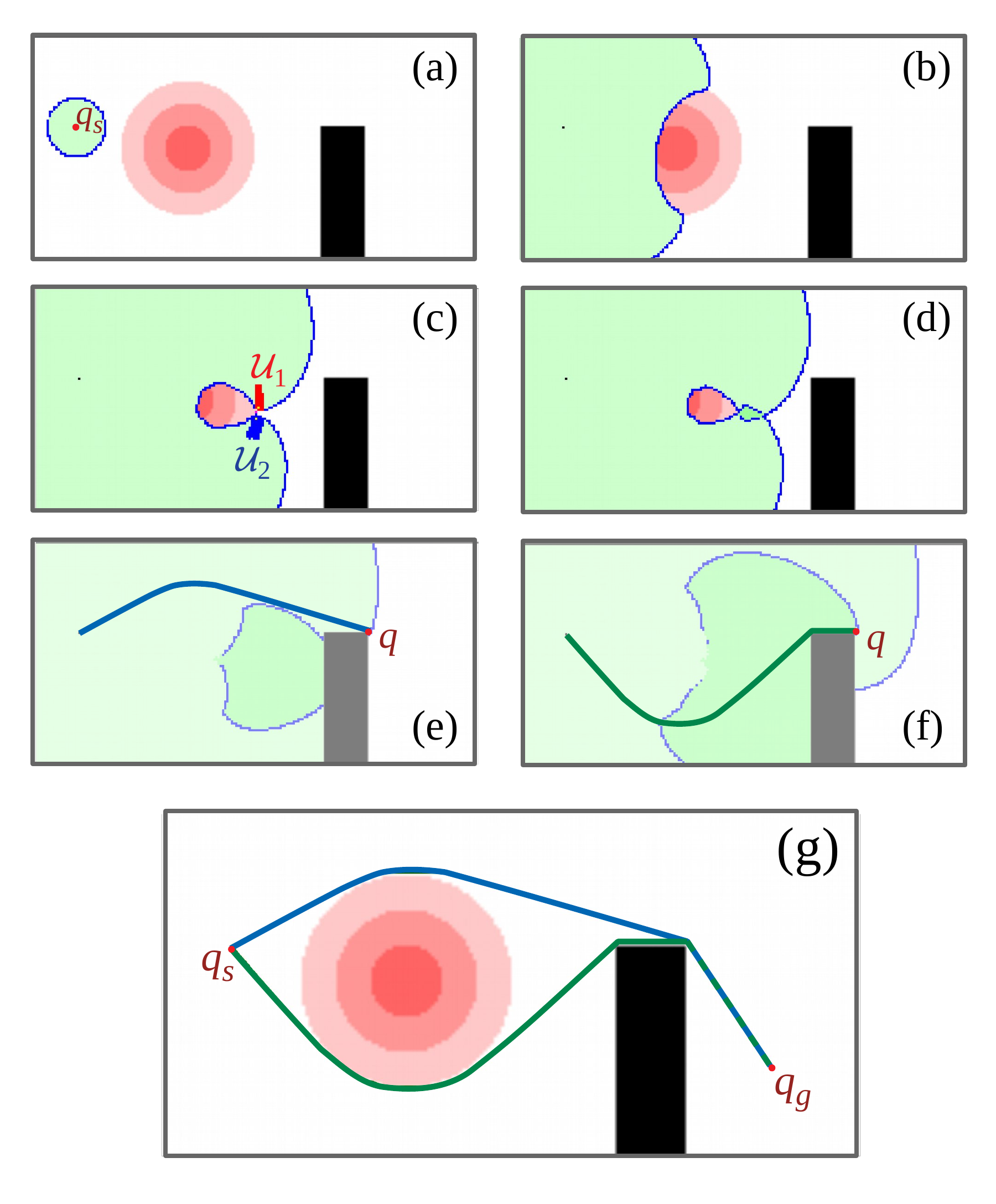}
    \caption{Overview of the proposed \emph{topo-geometric path planning} using search in \emph{neighborhood-augmented graph}: \changedAS{(a-b) Search propagates as a uniform wave-front before encountering any obstacle or high-cost region. (b-c) Two branches of the search emerge as the search wave-front is disturbed by the increased costs. (c) When two branches meet, paths leading up to the same configuration point will have disjoint path neighborhood sets ($\mathcal{U}_1$ and $\mathcal{U}_2$). (d) Branches stay separate as a result. Two topo-geometrically distinct paths to $q$ are found as (e) the first branch reaches the point $q$, (f) followed by the second branch. (g) Eventually the branches reach $q_g$ and two topo-geometrically distinct paths are obtained.}}
    \label{fig:tangent_test}
\end{figure}

\changedSB{
In a wave-front propagation type algorithm\footnote{In practice, this is Dijkstra's or S* algorithm on a discrete graph representation of the configuration space as will be discussed in the next section}, starting from $q_s$, 
this definition
is particularly relevant for identifying topo-geometrically distinct paths to the point where different \emph{branches} of the wave-front meet for the first time around an obstacle and/or high-curvature/cost regions (see Figure \ref{fig:tangent_test}(a-c)).
Such a point lies on the \emph{cut locus} of $q_s$~\cite{coverage:riemannian:IJRR:13}, and identification of topo-geometrically distinct paths to it based on distinct path neighborhoods (Figure \ref{fig:tangent_test}(c)) is critical in being able to identify, create and maintain distinct branches of the wave-front.
Once identified, the branches can be kept separated by distinguishing between (and identifying them as distinct) copies of the same configuration point reached from the different topo-geometrically distinct branches (Figure \ref{fig:tangent_test}(d-f)).
Subsequently, configuration points reached via the topo-geometrically distinct branches have different path neighborhoods (on the different branches) and hence can be identified as distinct with topo-geometrically distinct paths leading to them
(see Figure~\ref{fig:tangent_test}(e-g)).

}

\textbf{Path Neighborhoods:}
At this point we do not explicitly define or restrict the shape or size of the \emph{path neighborhoods}. The effect of parameters determining the shape and size of these neighborhoods will be discussed later, and in general will constitute tunable parameters that determine the performance of the algorithm (in terms of thresholds on size and shape of obstacles or high-curvature/-cost regions that give rise to the multiple topo-geometrically distinct paths).

}

\changedAS{
%

}

\section{Algorithm Development}
\input{sections/algorithm_development.tex}

\section{Modified Algorithm for Low Curvature/Cost Environments} \label{sec:merge_points}
\input{sections/merge_points_TBD.tex}

\section{Length Constrained Search \changedSB{for a Tethered Spatial Robot}} \label{sec:length-constrained-search}
\input{sections/length_constrained_search.tex}

\section{Conclusions}
\input{sections/discussion_conclusion}

\section*{Acknowledgments}

This material is based upon work supported by the National Science Foundation under Grant No. CCF-2144246.

\bibliographystyle{IEEEtran}
\bibliography{references}

\changedASnew{
\begin{appendix}
    \textbf{Proof of Proposition~\ref{proposition}:} We prove the equivalent contrapositive statement: \emph{If $\mathcal{U}_1 \cap \mathcal{U}_2 \neq \emptyset$, then $\text{sep}(\gamma_1',\gamma_2') \leq \frac{4 r_n}{1-\omega}$.}
    
    Suppose $u \in \mathcal{U}_1 \cap \mathcal{U}_2$ as shown in Figure~\ref{fig:proof}. Since the path neighborhood sets are constructed using A* search with radius $r_n$ and with the g-score of the primary search as the heuristic function with some heuristic weight $\omega$, for any $u \in \mathcal{U}_i$ we have 
    \begin{equation} \label{eq:u_def}
        d(u,q) + \omega d(u,q_s) \leq r_n + \omega (r_s - r_n)
    \end{equation}
    and
    \begin{equation}
        d(u,q) \leq r_n
    \end{equation}
    
    Since $\gamma_i' = \gamma_i \cap \mathcal{U}_i$ (given by the green and red colored lines in Figure~\ref{fig:proof}) and $\gamma_i$ (given by the black colored lines extending up to $q_s$) is the shortest path connecting $q$ in the open list and $q_s$, for any $v \in \gamma_i'$ we have
    \begin{equation}\label{eq:gamma'_def}
        d(v,q_s)+d(v,q) = r_s
    \end{equation}

    Substituting Equation~\ref{eq:gamma'_def} into Inequality \ref{eq:u_def},
    \begin{equation} \label{eq:main_subs}
    \begin{aligned}
        d(u,q) + \omega d(u,q_s) &\leq r_n + \omega (d(v,q_s) + d(v,q) - r_n)\\
        d(u,q) + d(v,q) & \leq r_n(1-\omega) + (1+\omega)d(v,q) \\
        &\quad + \omega(d(v,q_s)-d(u,q_s))
    \end{aligned}
    \end{equation}
    From the definition of $\gamma_i'$,
    \begin{equation}
        d(v,q) \leq r_n
    \end{equation}
    Using the following triangle inequality,
    \begin{equation}
        |d(v,q_s)-d(u,q_s)| \leq d(v,u) \leq d(u,q) + d(v,q)
    \end{equation}
    Inequality~\ref{eq:main_subs} reduces to
    \begin{equation}
    \begin{aligned}
        d(v,u) & \leq r_n(1-\omega) + (1+\omega)r_n + \omega d(v,u)\\
        (1-\omega) d(v,u) & \leq 2r_n
    \end{aligned}
    \end{equation}
    For $\omega \in [0,1)$,
    \begin{equation}
    \begin{aligned}
        sep(\gamma_1',\gamma_2') &\leq d(v_1,v_2) \\
                                &\leq d(v_1,u) + d(v_2,u) \\
                                &\leq 2d(v,u) \\
                                &\leq \frac{4r_n}{1-\omega}
    \end{aligned}
    \end{equation}

    \qed

\end{appendix}
}

\end{document}

%% file: sections/introduction.tex
\changedAS{Mobile robots primarily rely upon optimal path planning algorithms while navigating in complex environments. These robots could be operating on land, in air or underwater to perform transportation, exploration, coverage and even manipulation tasks. Among the most common approaches for computing optimal paths is discrete graph-search based methods~\cite{Ferguson_2008_6207,UrbanChallenge:VT:09,Ben:PR2:planning,conf-icra-SwaminathanPL15} that utilize search algorithms such as Dijkstra's, A* and its variants~\cite{dijkstra-dijkstra,Hart-Astar,Stentz-dstar,PRM:Kavraki:96,lavalle:rrt:2001,Daniel:JAIR:10,ijcai:CuiHG17,Bhattacharya:SStar:19}. These methods require the states of the robot and its configuration space to be abstracted as discretely represented graph at a desired resolution, on which the search algorithms can provide guarantees on the completeness and optimality depending on the type and resolution of the discrete representation.

There are robotics applications in which identifying paths from distinct homotopy classes is of particular interest. These applications include congestion reduction among a fleet of mobile robots, efficient exploration and coverage of unknown environments, navigation of tethered robots or cooperative manipulation using cable-connected robots~\cite{ICRA:14:tethered,cable:separation:IJRR:14,Wang:Cable-Controlled:2018,DARS:14:HRI, Kim:IROS13,wang2022congestion}. Topological path planning (TPP) methods address this interest by using \emph{augmented graphs} (\emph{homotopy} or \emph{homology}) to not only represent the configuration space, but also encode and keep track of topological classes of the paths taken to reach a state~\cite{cable:separation:IJRR:14,ICRA:14:tethered,planning:AURO:12,Homotopy-Planning:journal:18,wang2022congestion,Persistence-Plnning:TRO:15}. Research in this field primarily focuses on the design of homotopy invariants, which are quantities (elements from representations of the \emph{fundamental group} of the configuration space) that are computable as a function of a given path. Each vertex on the augmented graph can be identified with the homotopy invariant computed using the path leading up to the vertex. Then by deploying existing graph-search algorithms, it becomes possible to find a desired number of paths in different homotopy classes, each of which are locally-optimal in their respective homotopy classes.

Existing algorithms for topological path planning require non-trivial constructions in the configuration space. These constructions become increasingly complex when the configuration space is higher dimensional and of complex topology. This is mainly because topological invariants are computed based on some representation of the obstacles in the configuration space. For 2D configuration spaces with convex obstacles these representations could be as simple as some non-intersecting rays emanating from representative points. However, in higher dimensional spaces, obstacles of more complex topology and geometry may only be represented with skeletons and hyper-surfaces which are certainly more challenging to construct. Besides, in such spaces the fundamental groups are often not freely generated, and hence the computation of the homotopy invariants require complex equivalence check algorithms~\cite{Homotopy-Planning:journal:18}.

A task that cannot be accomplished at all by the existing topological path planning algorithms is the computation of multiple locally optimal paths that are homotopic but geometrically different. Certain robotics applications might entail configuration spaces in which there exists limited number of topological classes, and the existing algorithms, by nature, can only return a unique path in each distinct class. Some examples include mobile robots travelling on curved terrain, or aerial robots flying around prismatic obstacles (genus-0), where all the paths in the underlying configuration space belong to a single topological class. When the task is to efficiently explore around a high curvature landform (a hill or a valley) or navigate around obstacles while remaining tethered to a base, robots will benefit from being able to distinguish between locally optimal and geometrically distinct paths. Unlike topologically distinct paths, they can be continuously deformed into each other but only under significant increases and decreases in length (stretch and compression) during the deformation. \changedASnew{Algorithms utilizing Probabilistic Roadmap Planners (PRM) are used in \cite{jaillet2008pdr,zhou2021raptor} to compute paths that can only be transformed into each other under complex deformations. However, both methods require visibility information between points in the planning space to be able to classify paths.}

The topo-geometric path planning approach proposed in this paper enables computation of multiple locally optimal paths while remaining scalable to higher dimensional and complex configuration spaces. This is accomplished by constructing and comparing neighborhoods around locally optimal paths leading upto vertices on the search graph. Within every run of the search algorithm, we perform smaller sub-searches on the existing graph, starting from vertices in the open list. Every sub-search is performed for a fixed depth and the resulting subgraph constitutes the neighborhood for the vertex. By augmenting the graph with these neighborhood sets and using available graph-search algorithms, it becomes possible to find desired number of paths, each of which are locally optimal in the underlying configuration space without requiring any further constructions. These paths are guaranteed to be geometrically distinct, but they could also be topologically distinct depending on the topology of the configuration space.}

\changedAS{
The contributions of this paper are as follows:
\begin{itemize}
    \item Design of a novel topo-geometric path planning algorithm that uses a \emph{neighborhood augmented graph}, the construction of which is simple and does not require complex geometric constructions on the underlying configuration space \changedSB{(Sections~\ref{subsec:neighborhood_augmented_graph}, \ref{subsec:neighborhood_generation}). We also provide theoretical results on the algorithm (Section~\ref{sec:theory}).}
    \item Demonstration of the multi-class path planning capabilities of the algorithm in 2D and 3D configuration spaces of various topologies and cost functions (including cost functions and geometries that give rise to multiple locally optimal paths / geodesics even though they belong to the same topological class \changedSB{-- Section~\ref{sec:results-1}}).
    \item A further generalization/modification to the proposed method for computing locally-optimal paths in environments with extremely low curvature or cost variation that would otherwise not give rise to multiple \emph{branches} during the search in the neighborhood-augmented graph \changedSB{(Section~\ref{sec:merge_points})}.
    \item Implementation of the algorithm in path planning for a tethered robot navigating in 3D and with a tether-length constraint \changedSB{(Section~\ref{sec:length-constrained-search}), along with simulations and real robot experiments in different environments (Section~\ref{sec:length-constrained-search-simulation}, \ref{sec:length-constrained-search-experiment})}.
\end{itemize}
}



\begin{figure*}
    \centering
    \begin{subfigure}[b]{0.25\linewidth}
         \centering
         \includegraphics[width=\textwidth]{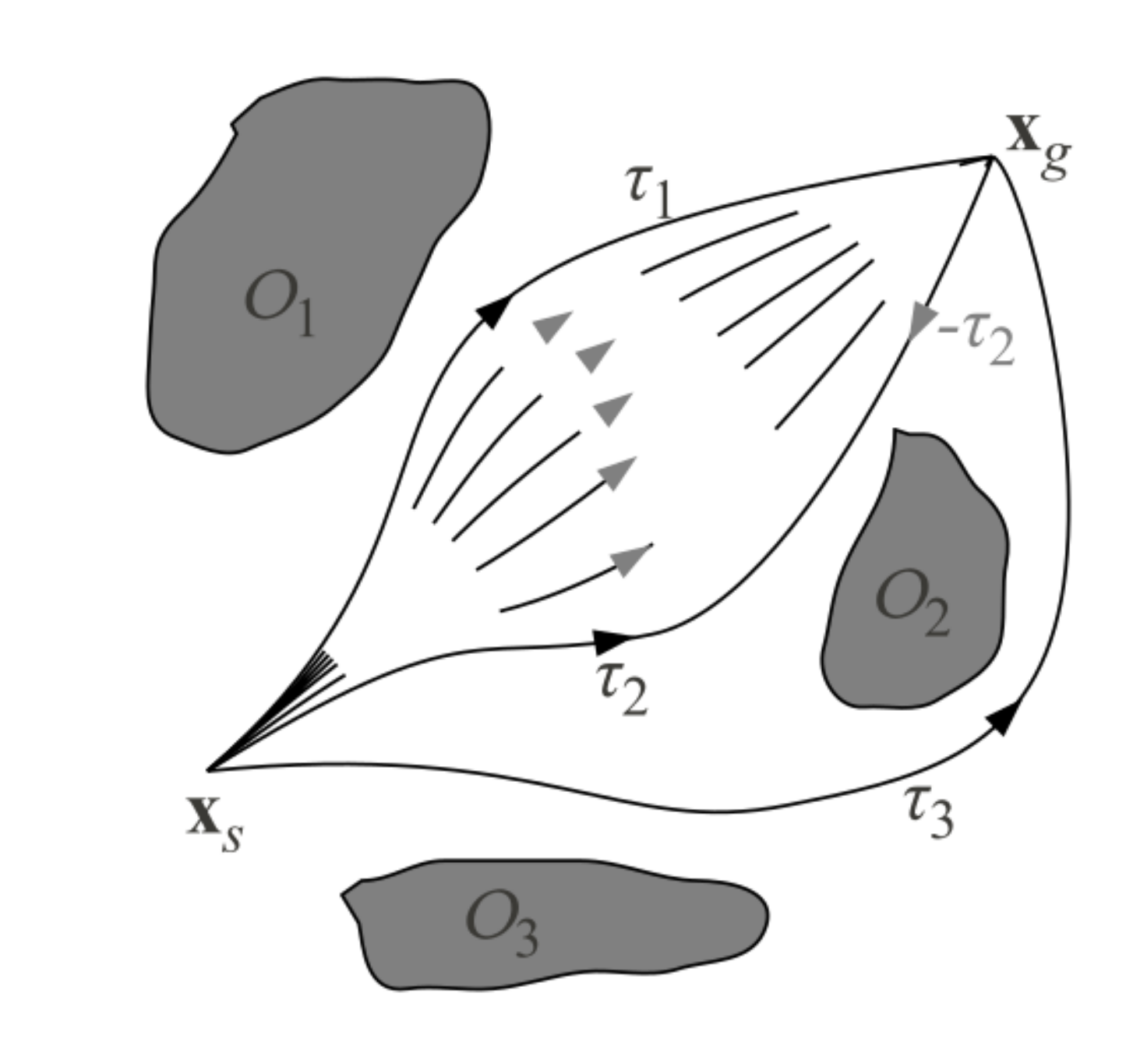}
         \caption{\changedAS{Homotopy equivalences: $\tau_1$ and $\tau_2$ are homotopic, while $\tau_3$ belongs to a different homotopy class.}}
         \label{fig:homotopy_def}
     \end{subfigure}
     \hfill
     \begin{subfigure}[b]{0.25\linewidth}
         \centering
         \includegraphics[width=\textwidth]{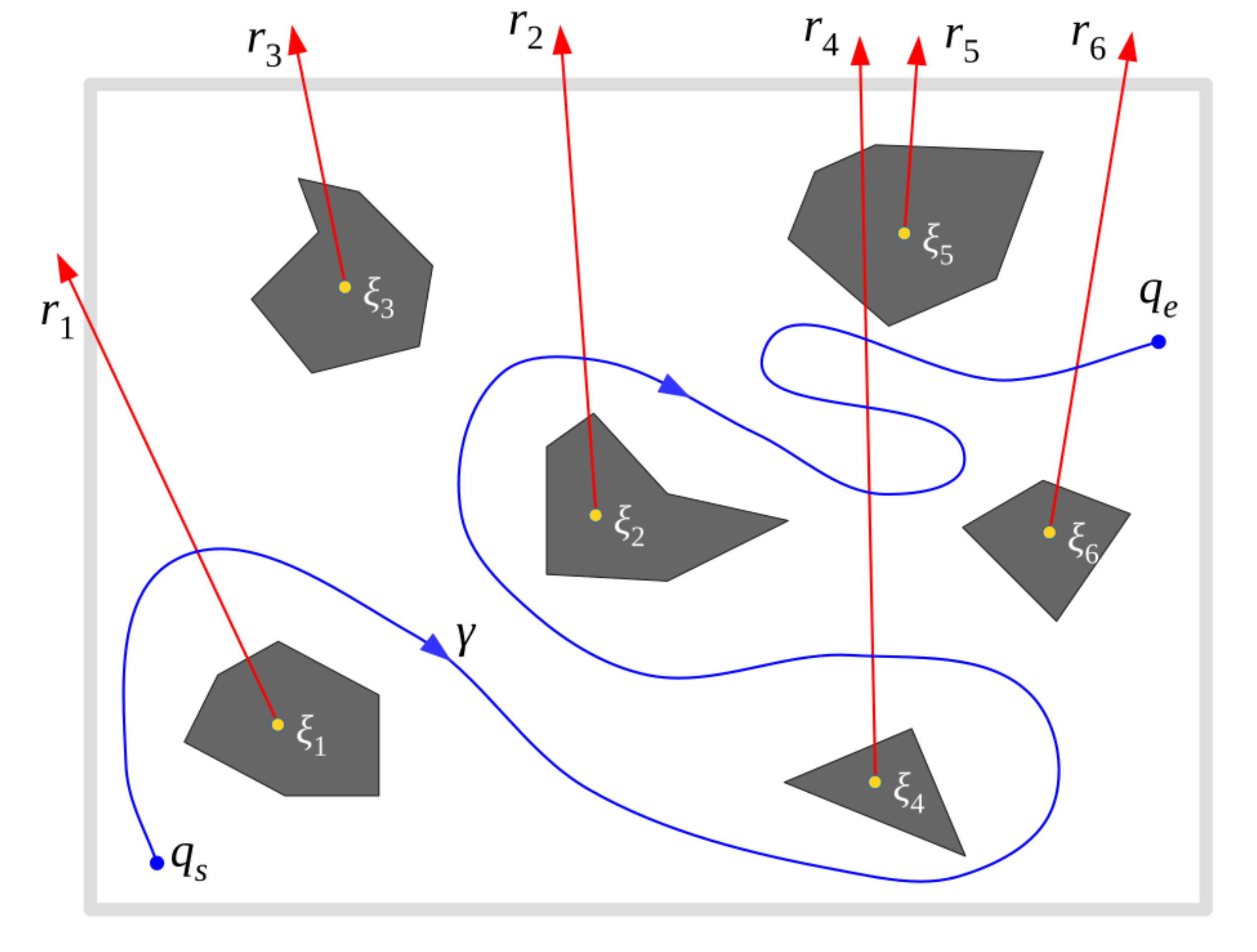}
         \caption{In 2D environments, rays emanating from representative points on the obstacles can be used to generate word-based homotopy invariant called $h$-signatures~\cite{ICRA:14:tethered,cable:separation:IJRR:14}.}
         \label{fig:word_based_signature}
     \end{subfigure}
     \hfill
     \begin{subfigure}[b]{0.45\linewidth}
         \centering
         \includegraphics[width=\textwidth]{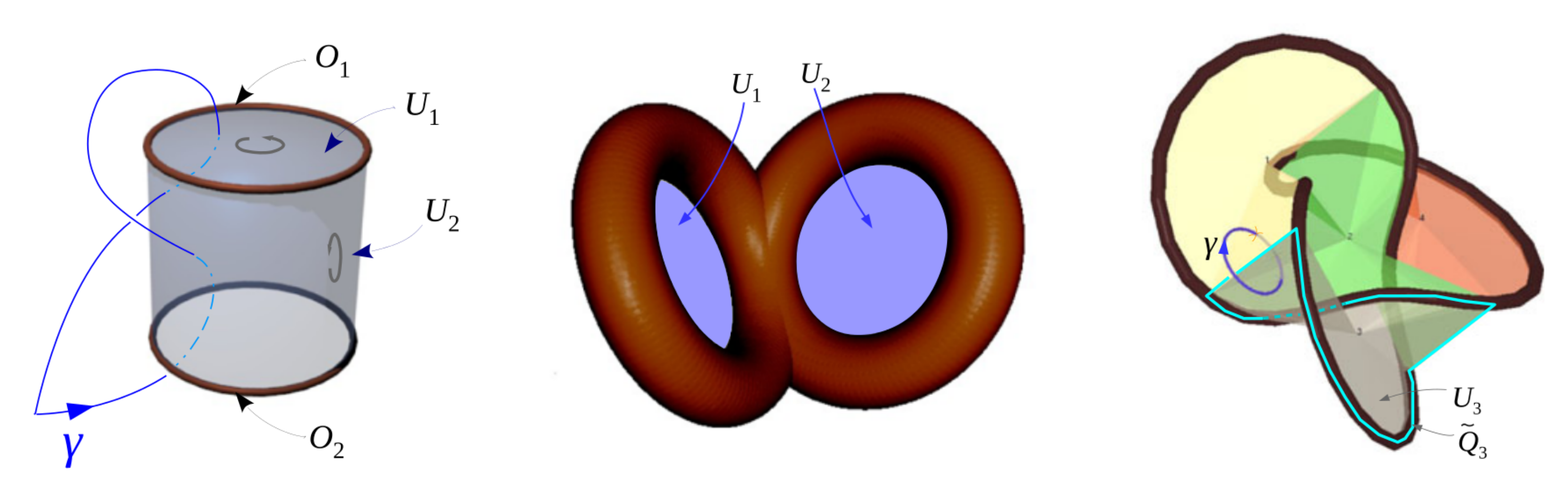}
         \caption{In 3D environments, rays need to be replaced by surfaces bounded by obstacles with genus higher than zero. Such constructions are significantly more difficult,
                  and the $h$-signature computation is highly nontrivial especially in presence of knotted or linked obstacles~\cite{Homotopy-Planning:journal:18}.}
         \label{fig:surfaces}
     \end{subfigure}
     \begin{subfigure}[b]{0.25\linewidth}
         \centering
         \includegraphics[width=\textwidth]{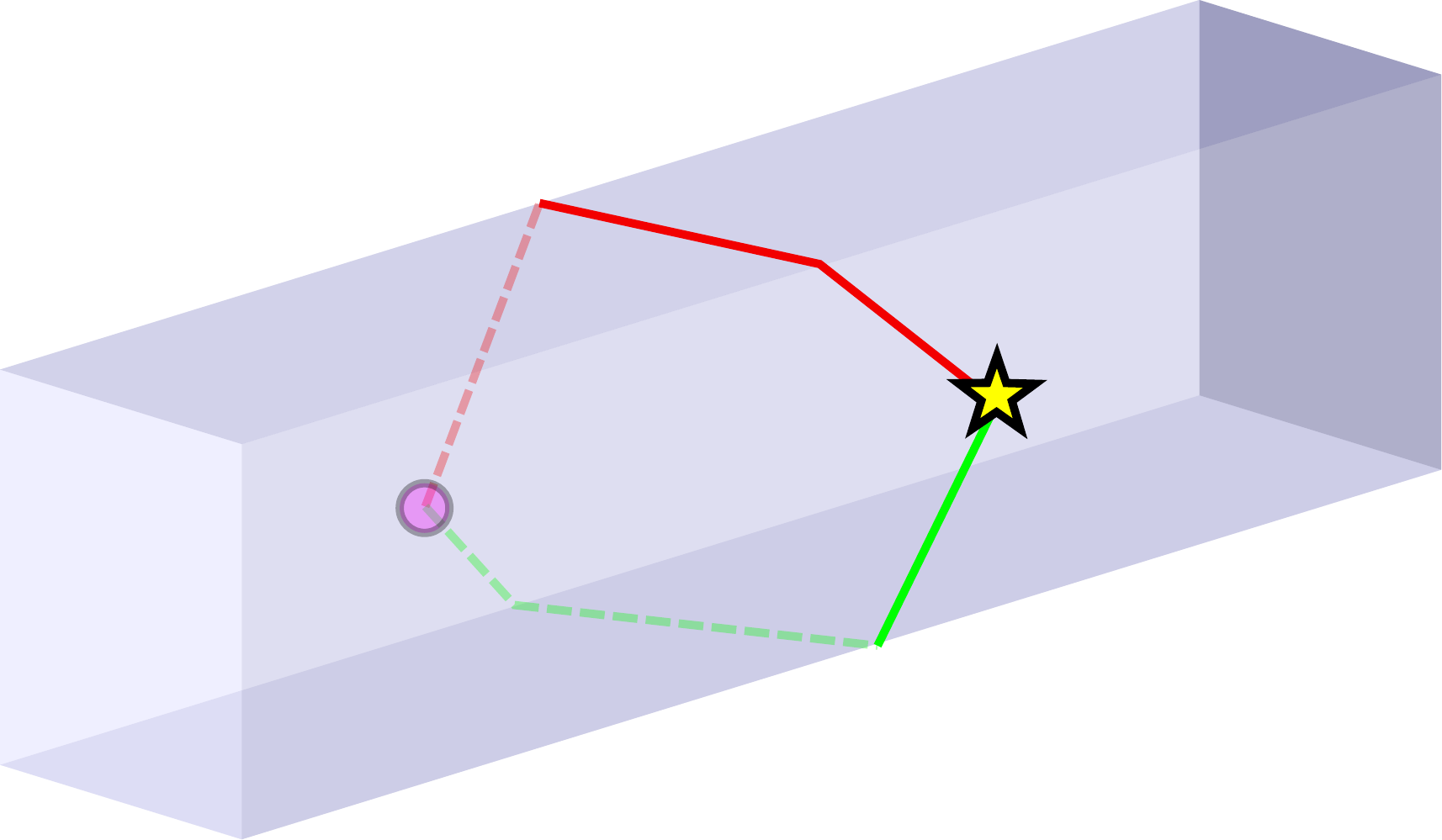}
         \caption{\changedAS{There exist multiple, locally shortest paths around a rectangular prism, even though these paths are topologically equivalent (belong to the same homotopy class).}}
         \label{fig:long_rectangle}
     \end{subfigure}
     \hfill
     \begin{subfigure}[b]{0.15\linewidth}
         \centering
         \includegraphics[width=\textwidth]{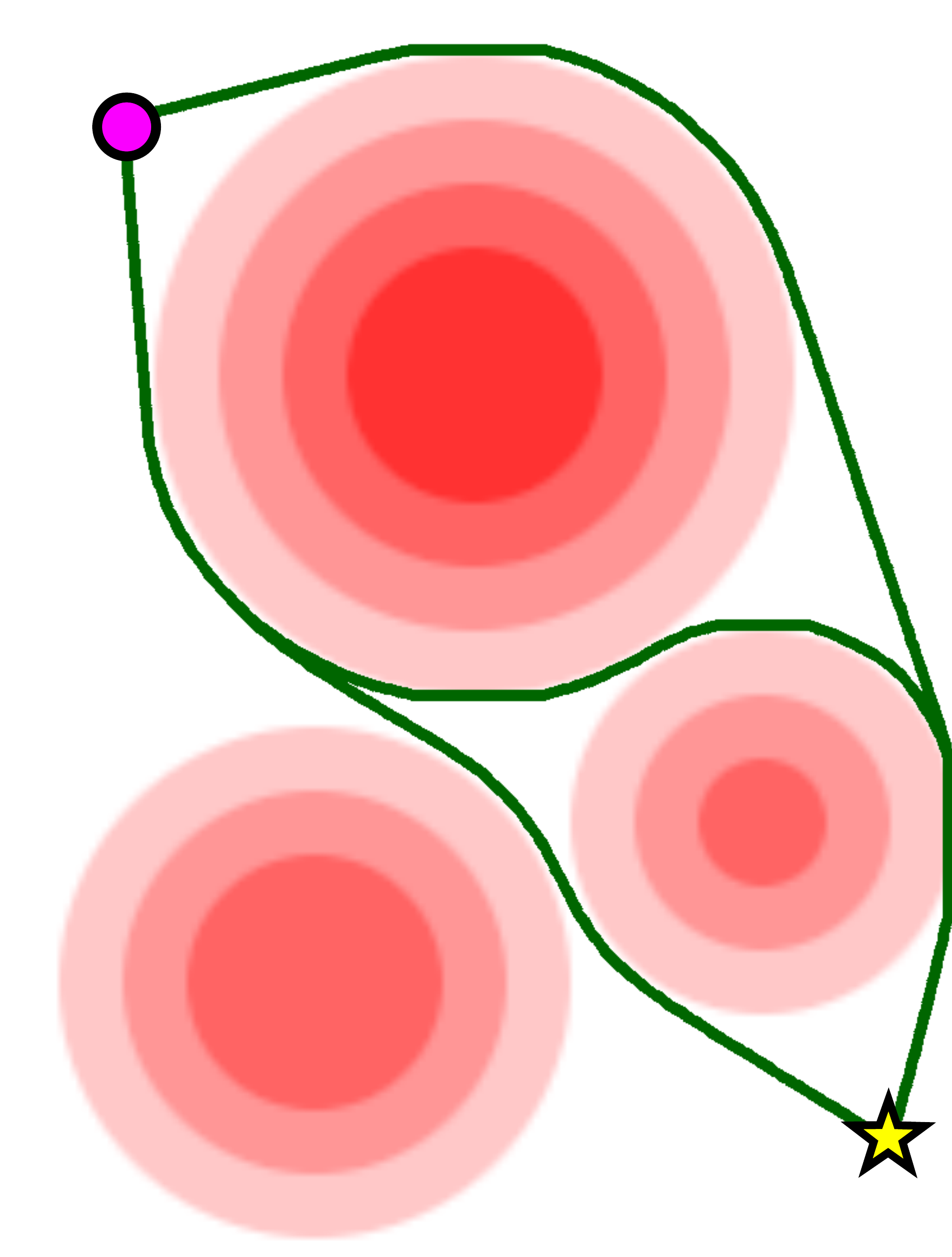}
         \caption{\changedAS{On a 2D environment with high cost regions, there exists multiple, locally shortest paths that are topologically equivalent.}}
         \label{fig:nonuniform_1}
     \end{subfigure}
     \hfill
     \begin{subfigure}[b]{0.25\linewidth}
         \centering
         \includegraphics[width=\textwidth]{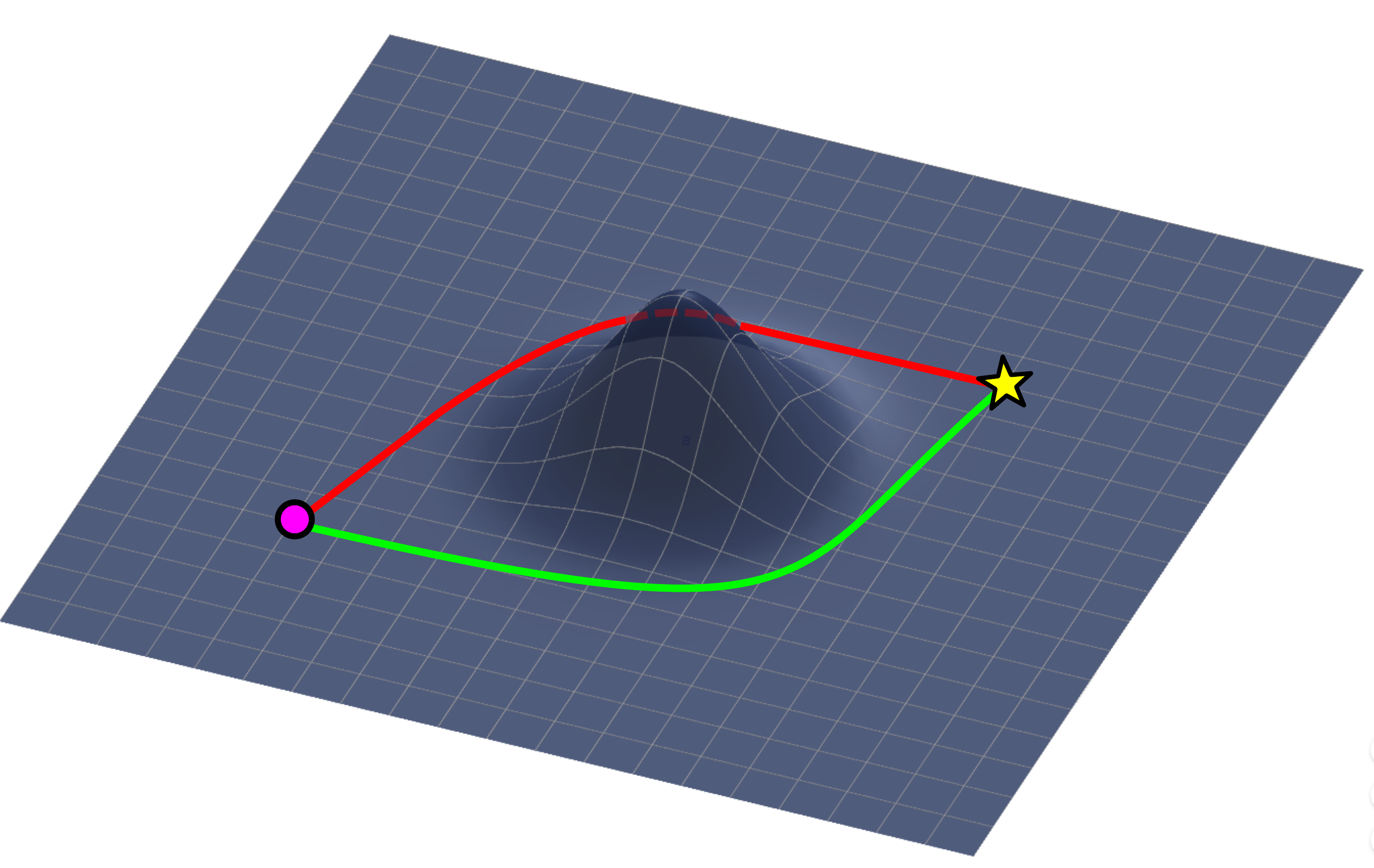}
         \caption{\changedAS{On a surface with non-zero curvature, there exists multiple, locally shortest paths that are topologically equivalent (can be continuously deformed into one another although it requires a path to be \emph{stretched} over the \emph{hill}).}}
         \label{fig:nonuniform_2}
     \end{subfigure}
     \hfill
     \begin{subfigure}[b]{0.3\linewidth}
         \centering
         \includegraphics[width=\textwidth]{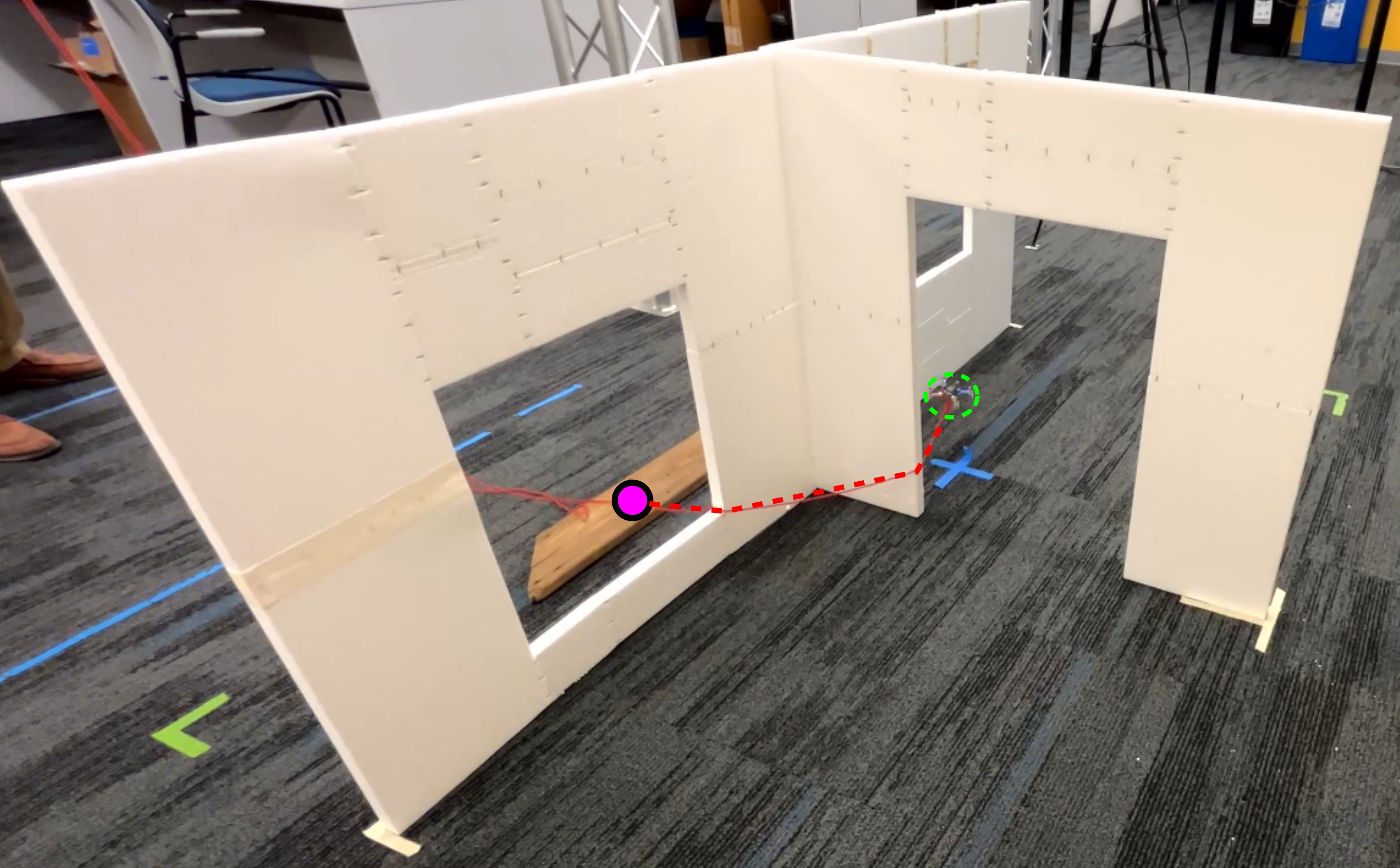}
         \caption{\changedAS{A tethered quadrotor performing a navigation task in a building-like environment with doors/windows need to reason about topo-geometrically distinct paths.}}
         \label{fig:rr_exp}
     \end{subfigure}
        \caption{\textbf{Background and Motivation --} \changedAS{\emph{Top row} (a-c): Existing/prior work on \emph{topological path planning (TPP)} for computing optimal paths in different homotopy classes. The constructions and computations gets significantly harder in 3-D configuration spaces, requiring complex group-theoretic reasoning in presence of obstacles that are knotted or linked (such structures are encountered in urban structures such as buildings with stairways, doors and windows).
        \emph{Bottom row} (d-g): Motivation behind current work -- Even in absence of topologically-distinct classes of paths, there can exists different geometrically-distinct locally optimal paths (d-e). This, along with the increased complexity of TPP in 3D (c), motivates our current topo-geometric planning algorithm that creates an unified framework for computing multiple locally-optimal paths connecting a start and a goal configuration.}}
        \label{fig:main_figure}
\end{figure*}

%% file: sections/algorithm_development.tex


\subsection{Preliminaries: Graph Search-based Planning}

\subsubsection{Discrete Graph Representation of a Configuration Space for Shortest Path Planning}

\changedSB{
In traditional graph-search based shortest path-planning in a configuration space, a graph $G = (V,E)$, is constructed as a discrete representation the configuration space, where the vertex set consist of points sampled (uniformly or in a randomized manner) from the valid configuration space.
A vertex, $q\in V$, is typically represented by its spatial coordinates, and an edge $e=(q_1,q_2)\in E\subseteq V\times V$ connects neighboring vertices $q_1$ and $q_2$.
Search-based algorithms construct and explore the graph in an incremental manner: Starting at a start vertex, $q_s\in V$, it uses a wavefront propagation type approach, where it maintains an \emph{open list} (the \emph{exploration front}) and generates new neighboring vertices of the open list as it expands the \emph{front}. This process continues until a desired goal vertex, $q_g\in V$, is reached.
While there is a vast variety of search based algorithms -- some purely for finding shortest paths in graphs (such as Dijkstra's, A*~\cite{dijkstra-dijkstra,Hart-Astar}), some using line-of-sight type information for finding paths that are closer to optimal in the underlying configuration space (Theta*~\cite{Daniel:JAIR:10}), and others construct simplicial complexes from the graph for achieving the same without line-of-sight information (such as S*~\cite{Bhattacharya:SStar:19}) -- at their core they require a \emph{neighbor or successor function}, $\mathcal{N}_G$, with the following properties:
\begin{enumerate}
    \item[i.] Given a vertex $q\in V$, $\mathcal{N}_G(q)$ returns the \emph{neighbors} (vertices that are connected to $q$ by edges) of the vertex (and the cost of the edges connecting them);
    \item[ii.] The algorithm also needs to be able to tell if a vertex returned by the neighbor function is something already explored/encountered in the past. For the usual discrete graph representation of a configuration space, this is usually as simple as comparing the coordinates of the new vertex with the coordinates of the existing/explored vertices.
\end{enumerate}
}

\begin{figure}[h!]
    \centering
    \includegraphics[width=0.6\linewidth]{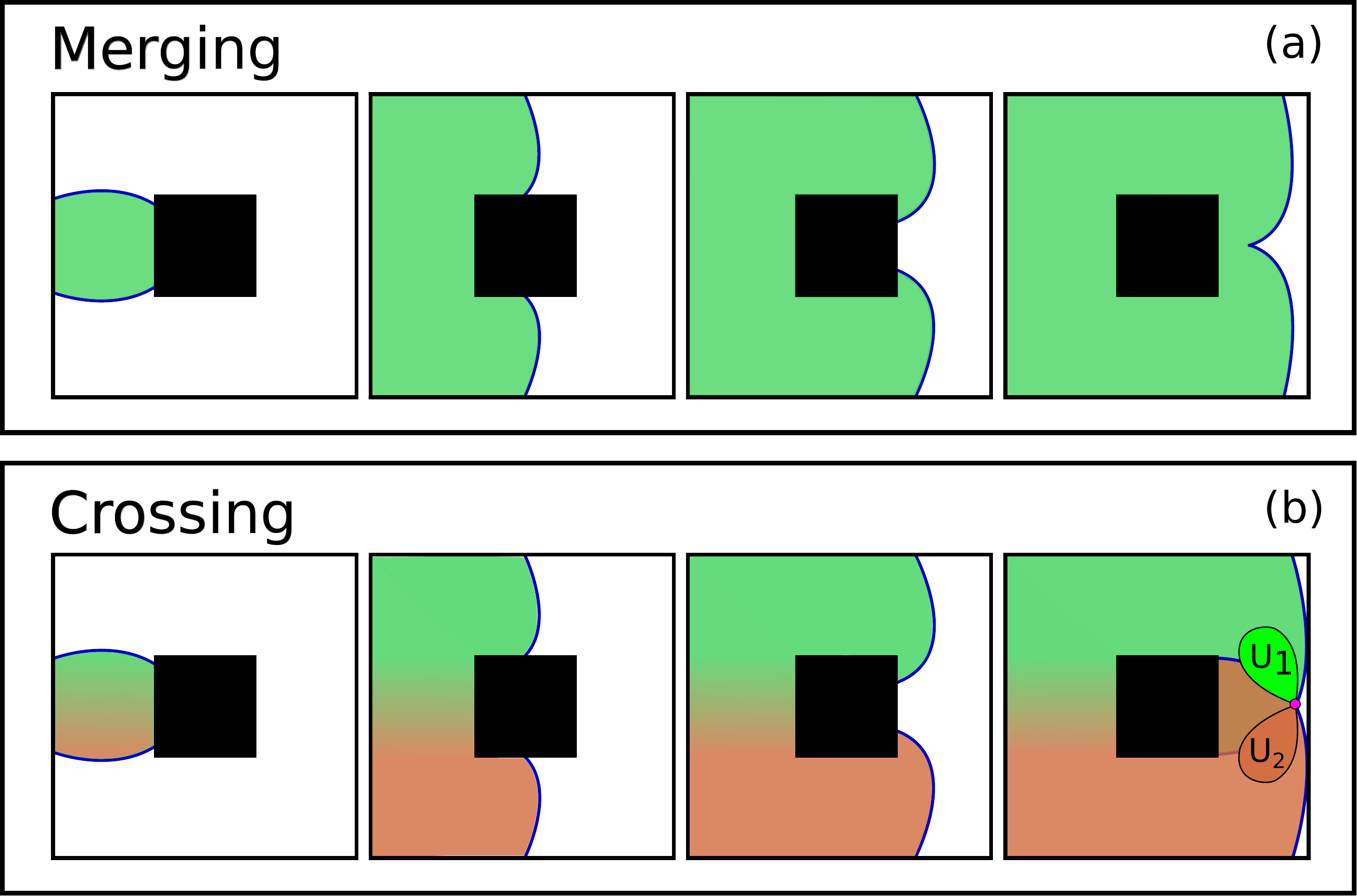}
    \caption{Illustration of search progress in $G$ vs. $G_N$: (a) Stages of search on a regular (non-augmented) discrete graph, $G$, representation of the configuration space demonstrates how different branches emerge and merge due to an obstacle in the environment. (b) Stages of search on a neighborhood-augmented graph demonstrates how different branches emerge and stay separated (different colors have been used to indicate different branches that remain separated). When a point, $q$, can be reached via different branches, corresponding path neighborhood sets ($\mathcal{U}_1$ and $\mathcal{U}_2$) are disjoint, and hence are represented by two distinct vertices, $(q,\mathcal{U}_1),(q,\mathcal{U}_2)\in V_N$. Two distinct paths to $q$ are obtained by reconstructing path from $(q,\mathcal{U}_1)$ and $(q,\mathcal{U}_2)$ in $G_N$.}
    \label{fig:merging_vs_separate}
\end{figure}

On a graph that represents an obstacle-free Euclidean space, the search will propagate as an uniform wave. The exact shape of the search wave will depend on the specific search algorithm and the heuristic (if any being used). In case of planning in spaces with obstacles or nonuniform path costs 
these artifacts will obstruct or slow down the propagation of the search in certain directions as shown in Figure~\ref{fig:merging_vs_separate}(a). 
As a result, multiple branches of the search wave will emerge, following through alternative paths around the artifact. However, when two branches meet at the downstream of the artifact (\textit{downstream} of the artifact refers to the set of vertices that are further away from the start vertex compared to the artifact), they merge into one, as they pass through vertices with the same coordinates. Therefore, the traditional approach does not allow any distinction between paths, or even multiple paths to be found to the same vertex.

\subsubsection{Augmented Graph for Multi-class Path Planning}

\changedSB{In order to find multiple paths to the same vertex in the configuration space using a search-based approach, an \emph{augmented graph}, $G_A = (V_A, E_A)$, needs to be constructed from $G$. Vertices in $V_A$ are of the form $(q,\alpha)$ for $q\in V$ and $\alpha$ an invariant of a path class (\emph{i.e.} a value/quantity -- numeric or otherwise -- that uniquely identifies the path class). Thus, corresponding to a vertex $q\in V$ there are multiple vertices of the form $(q,\alpha_1), (q,\alpha_2), \cdots \in V_A$, each corresponding to the vertex $q$ reached via a different path class identified by invariants $\alpha_1,\alpha_2,\cdots$.

\emph{{An Example -- $h$-augmented Graph:}}
The idea is most easily demonstrated by the construction of $h$-augmented graph in context of topological path planning in multiple homotopy classes~\cite{ICRA:14:tethered,cable:separation:IJRR:14,Wang:Cable-Controlled:2018}.
The idea behind construction of the $h$-augmented graph, $G_h=(V_h,E_h)$, is to append each vertex of $G$ with the homotopy invariant of a path (called the $h$-signature, which is represented by a ``word'' constructed by concatenating the letters corresponding to rays emanating from obstacles that the path crosses -- Figure~\ref{fig:main_figure}(b)) leading up to that vertex from a given start vertex $q_s\in V$.
With the $h$-signature associated with the start vertex being the empty word, $\ldq~\rdq$, the construction of this graph can be made in an incremental manner and allows the algorithm to keep track of the homotopy class of the paths leading to a vertex during the execution of a search (Figure~\ref{fig:h-aug} inset). This, in effect, allows us to distinguish a vertex based on the homotopy class of the path taken to reach it from the start -- two branches of the search front (open list) in two different homotopy classes leading to the same vertex $q_g$ in $V$ actually leads to two distinct vertices in $V_h$ (Figure~\ref{fig:h-aug}).
\begin{figure}
	\centering
		\includegraphics[width=0.46\linewidth, trim=0 0 0 0, clip=true]{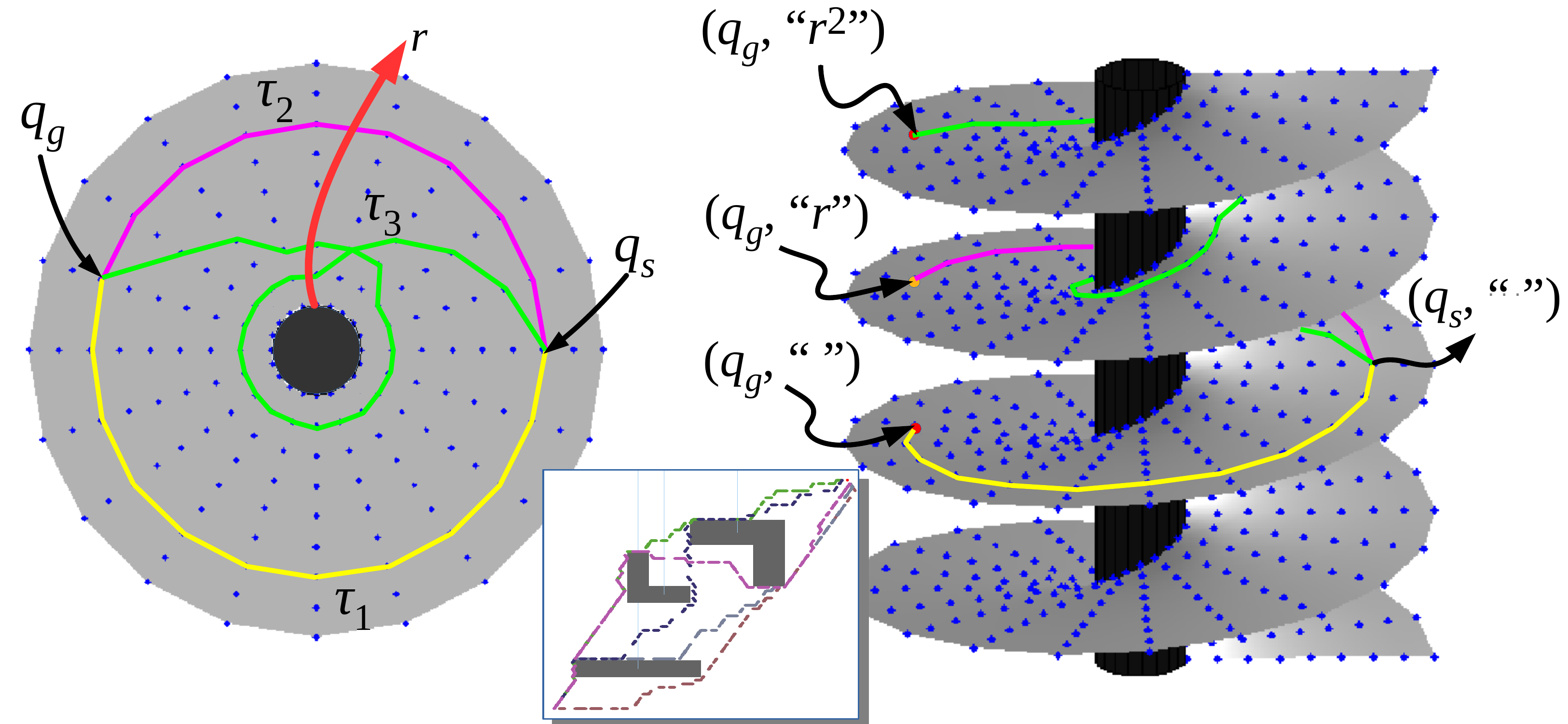} 
	\caption{Visualization of the $h$-augmented graph~\cite{ICRA:14:tethered,cable:separation:IJRR:14,Wang:Cable-Controlled:2018}: \emph{Left:} A planar configuration space (light gray), with a single obstacle (black), and the vertex set, $V$ (blue dots), in it. \emph{Right:} The vertex set, $V_h$. Note how the \changedSB{paths} \emph{lift} to have different goal points of the form $(q_g,*)\in V_h$ corresponding to the same goal $q_g\in V$. \emph{Inset:} Optimal paths in different homotopy classes in an environment with multiple obstacles.} \label{fig:h-aug}
\end{figure}
}

\changedAS{
\emph{{Shortcomings of Multi-class Planning using $h$-augmented Graph:}}
One common challenge in computing topological invariants such as $h$-signature is that their computation requires geometric constructions based on a complete knowledge of obstacles in the environment (rays in the $2$-D case, for example -- Figure~\ref{fig:main_figure}(b)). For higher dimensional and complex configuration spaces, it becomes non-trivial to make analogous geometric constructions (Figure~\ref{fig:main_figure}(c)), and even if such constructions are possible, the computation of $h$-signature needs to consider equivalence of certain set of words~\cite{Bhattacharya2018}.
%
Furthermore, although topological path planning using $h$-augmented graph can efficiently compute multiple, locally optimal and topologically distinct paths, they are still unable to compute multiple paths within the same topological class that could be geometrically distinct. This scenario usually occurs when planning on a surface with non-zero curvature or in 3D spaces with genus-0 obstacles (Figure~\ref{fig:main_figure}(d-f)).
}

\subsection{Neighborhood-augmented Graph}\label{subsec:neighborhood_augmented_graph}

\changedSB{In order to do away with complex geometric constructions in topological path planning, and to be able to find locally-optimal paths that are topo-geometrically distinct (Definition~\ref{def:topo-geometric}), in this paper we propose a \emph{neighborhood-augmented graph}, $G_N = (V_N, E_N)$, that is constructed from the given discrete graph representation of the underlying configuration space, $G = (V,E)$, by augmenting each vertex with a ``\emph{path neighborhood}'' that is representative of the tangent vector of the path leading to the vertex (Figure~\ref{fig:top_geo}). Just like the previously-described augmented graphs, corresponding to a vertex $q\in V$, there can exist multiple distinct vertices, $(q,\mathcal{U}_1), (q,\mathcal{U}_2), \cdots \in V_N$, where \underline{each $\mathcal{U}_i$ consist of \textbf{references/pointers to} vertices in $V_N$} (\emph{i.e.}, $\mathcal{U}_i \subset V_N$) that make up the neighborhood of a portion of the path leading the vertex $(q,\mathcal{U}_i)$. We refer to each such $\mathcal{U}_i$ as a \emph{path neighborhood set} corresponding to the path leading to the vertex $v$.}

\changedAS{
For topo-geometrically distinct paths, the path neighborhood sets are expected to be disjoint as illustrated in Figures~\ref{fig:top_geo} \& \ref{fig:merging_vs_separate}(b).
\changedSB{While, if a vertex is reached 
via paths that are topo-geometrically similar
(\emph{e.g.}, from a different parent/vertex), there will be significant overlap between the path neighborhood sets. Thus, we define
\begin{definition}[Equivalence of Vertices in $G_N$]
    Given $(q,\mathcal{U}), (q',\mathcal{U}')\in V_N$,
    they are called \emph{equivalent} ({i.e.}, considered to be the same vertex, and denoted as $(q,\mathcal{U}) \equiv (q',\mathcal{U}')$), if
    $q=q'$ and $\mathcal{U} \cap \mathcal{U}' \neq \emptyset$.
\end{definition}
}
This comparison of vertices is critical in incremental construction of the graph, $G_N$ -- every time a new vertex is generated, it can be compared against already-generated/existing vertices to determine if it is a new vertex or if it is the same vertex as an existing one.}

\emph{An Implementational Detail:}
It is to be noted that for two path neighborhood sets to intersect (\emph{i.e.}, for the computation of $\mathcal{U} \cap \mathcal{U}'$), they must contain a common vertex that not only has the same coordinates/configuration but also the same path neighborhood sets \changedSB{(that is, $\mathcal{U} \cap \mathcal{U}' \neq \emptyset ~\iff~ \exists~ (w,\mathcal{W})\in \mathcal{U},~ (w',\mathcal{W}')\in \mathcal{U}' ~\text{s.t.}~ (w,\mathcal{W}) \equiv (w',\mathcal{W}')$).
At a first glance it may appear that the computation of equivalence thus requires an iterative computation of other equivalences for vertices in the path neighborhood sets. However, in implementation, since path neighborhood sets are stored as \emph{pointers} to vertices in $V_N$, it is sufficient to check if the sets of pointers, $\mathcal{U}$ and $\mathcal{U}'$, have any common pointer. If the pointers are the same, that will automatically imply that they have the same (hence overlapping) path neighborhood sets (since otherwise they would have been identified as different vertices in the first place and would have been assigned different pointers).}

\changedSB{\emph{Neighbor/Successor Function:}}
For any vertex $(q,\mathcal{U}) \in V_N$, 
\changedSB{the neighbor function of the configuration graph, $G$, returns the configuration of (which are usually represented by coordinates in the configuration space) vertices that are adjacent to $q$. Suppose} $\{q_1, q_2,\cdots\} = \mathcal{N}_G(q)$.
\changedSB{The path neighborhood sets of each $q_i$ is then computed by actually computing a neighborhood (by collecting pointers to vertices in $V_N$) of $(q,\mathcal{U})$ by performing a short secondary exploration/search in the current/existing $G_N$. The details of this secondary search is described in Section~\ref{subsec:neighborhood_generation}. If this secondary search returns a path neighborhood set $\mathcal{U}'$, the neighbor/successor function of $G_N$ can then be described as $\{(q_1,\mathcal{U}'), (q_2,\mathcal{U}'), \cdots \} = \mathcal{N}_{G_N}\left( (q,\mathcal{U}) \right)$}. The cost of each edge on $G_N$ is the same as its projection on the original graph $G$ \changedSB{(that is, $\mathcal{C}_{G_N}\left( (q,\mathcal{U}), (q_i,\mathcal{U}') \right) = \mathcal{C}_G(q,q')$)}.


\input{pseudocodes/pseudo_NAsearch.tex}

\input{pseudocodes/pseudo_stopSearchGoal.tex}

\changedSB{A search in the graph $G_N$ can be performed using any search algorithm.
\changedSB{Throughout this paper we have presented results from searches in $G_N$ performed using Dijkstra's search and S* search~\cite{Bhattacharya:SStar:19} for computing different topo-geometrically distinct paths that are optimal in the underlying configuration space.}

As a simple illustration, however, we present a}
pseudo-code for the search \changedSB{using Dijkstra's algorithm} on an incrementally constructed neighborhood-augmented graph in Algorithm~\ref{alg:NASearch}. \changedAS{The traditional way to use this algorithm is to insert the goal-based stopping criteria given in Algorithm~\ref{alg:stopSearchGoal} as the input, provided a goal coordinate and number of desired paths. However, depending on the application, there could be other uses of the algorithm, utilizing alternative stopping criteria, as it will be further discussed in Section~\ref{subsec:length_constrained_search}.}

\begin{figure*}[h!]
    \centering
    \includegraphics[width=\linewidth]{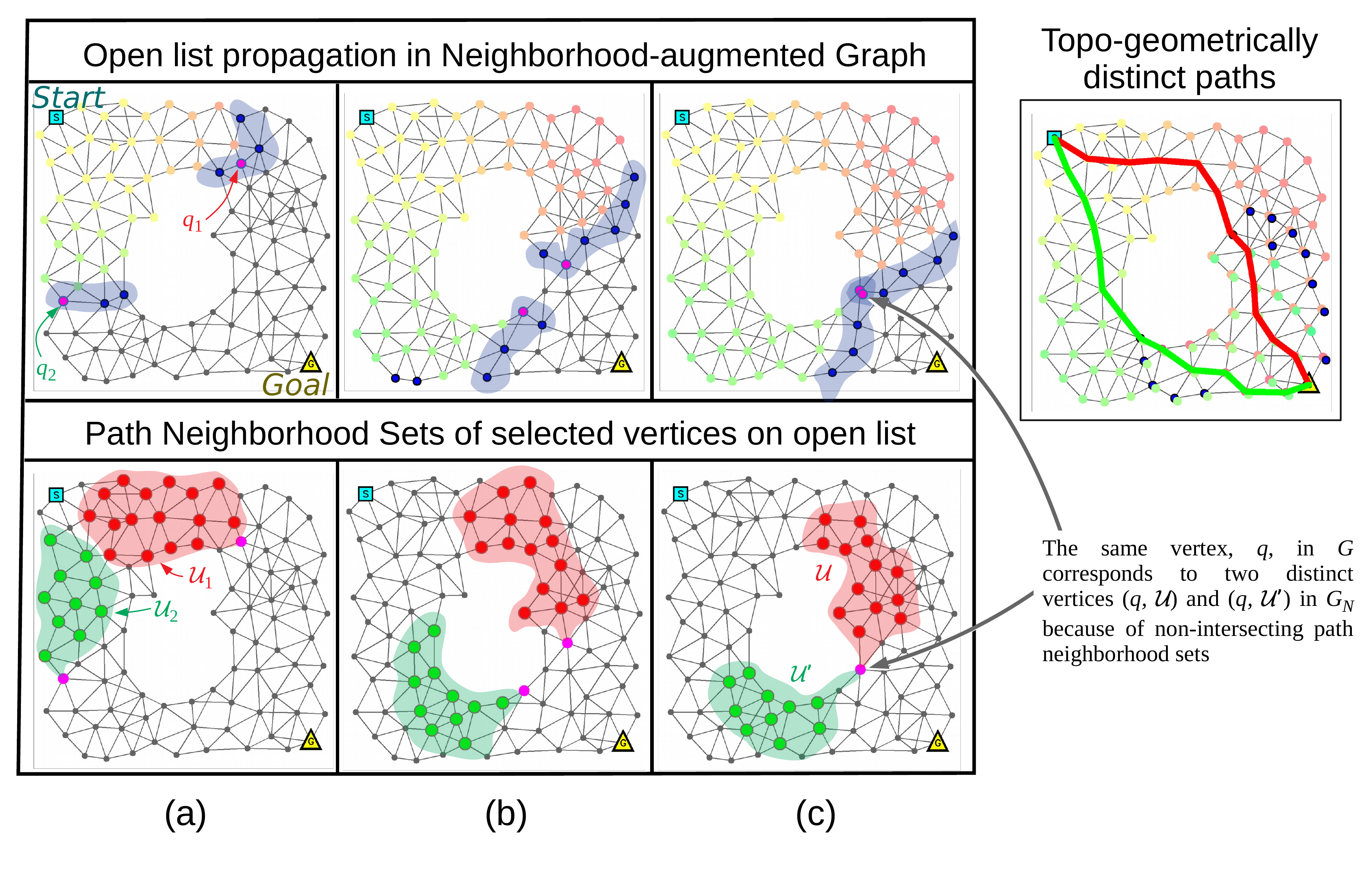}
    \caption{Path planning on \changedSB{neighborhood-augmented graph, $G_N$, using Dijkstra's search (actual \textbf{result}, with highlights and labeling for aiding explanation)}. The planar graph that is visible in the figures is the configuration graph $G = (V,E)$.
    The neighborhood-augmented graph is constructed incrementally, and the top row shows the vertices in the \emph{open list} at different instants during the progress of the search (colored/highlighted in blue).
    However, a vertex in the neighborhood-augmented graph, $G_N$, is not just a vertex in $G$, but also contains information about a path neighborhood set leading to the vertex (maintained as a list of pointers to already-generated vertices in $V_N$). This is shown in the second row, where two vertices, $(q_1,\mathcal{U}_1)$ and $(q_2,\mathcal{U}_2)$, in the two different \emph{branches} of the search (created due to the central \emph{hole}) in the graph, are chosen to illustrate their respective path neighborhood sets (shown/highlighted in red and green respectively).
    However, as demonstrated in column (c) of the figure, when the two branches \emph{meet}, corresponding to the same vertex $q$ in the configuration graph, we end up constructing two distinct vertices, $(q,\mathcal{U})$ and $(q,\mathcal{U}')$, in $G_N$.
    %
    Two locally shortest topo-geometrically distinct paths to the goal are found as a result (last column). The cost function used for this search example is the Euclidean distance between the vertices.
    }
    \label{fig:discrete_graph_combined}
\end{figure*}

\changedAS{
Figure~\ref{fig:discrete_graph_combined} illustrates the incremental construction of a neighborhood-augmented graph based on a given discrete graph.
Two separate search \emph{branches} emerge due to a hole in the underlying discrete graph.
Neighborhoods for selected vertices are shown in the bottom row. 
Vertices shown in the first \changedSB{two} columns are trivially distinct as they do not share the same coordinates. In the \changedSB{third} column, two vertices with the same coordinates (\emph{i.e.}, the same vertex in $G$) are selected for which the path neighborhood sets are disjoint as they are reached via different branches of the search. As a result, two copies of vertices (with two disjoint path neighborhood sets) are maintained separately within the neighborhood-augmented graph. Whenever a distinct vertex at the goal coordinates is expanded, it is possible to reconstruct a topo-geometrically distinct path from the start to the goal coordinate using the corresponding goal vertex and the constructed neighborhood-augmented graph with a \textit{path reconstruction} subroutine. Two of such paths are shown in the last column of Figure~\ref{fig:discrete_graph_combined}.

For given shape characteristics/geometry of the path neighborhood sets, and geometric artifacts of the configuration graph (influenced by holes, obstacles, high-cost or non-zero curvature regions), branches of the search around such artifacts may not merge, but stay separate (cross-over) and progress further, as there can be vertices at the same coordinates that are reached via significantly different paths and hence have different path neighborhood sets.
\changedSB{The effect of the size/geometry of the path neighborhood sets on the ability to find topo-geometrically distinct paths around obstacles of different size/geometry is discussed in more detail in Section~\ref{subsec:implementation_details}.}
Performing the search on the neighborhood-augmented graph computes alternative and locally optimal paths that are topo-geometrically different than the globally optimal path. An illustration of the path search using S* search algorithm in a discrete representation of a continuous configuration space using a neighborhood-augmented graph 
is shown in Figure~\ref{fig:merging_vs_separate}(b).
}


\subsection{Neighborhood Generation}\label{subsec:neighborhood_generation}

During the search in the neighborhood-augmented graph (Algorithm~\ref{alg:NASearch} -- also referred to as \emph{\textbf{primary search}} henceforth),
a path neighborhood set, $\mathcal{U}$, of a vertex $v\in V_N$ need to be computed (this consists of a set of vertices within some distance $r_n$ of $v$ in the neighborhood \changedSB{of the path leading to $v$ in the} neighborhood-augmented graph) -- Line~\ref{ln:computePNS} of Algorithm~\ref{alg:NASearch}. To compute this path neighborhood set, a \emph{secondary search} is performed on the current neighborhood-augmented graph, $G_N$ (using A* algorithm), starting from the vertex of interest $v$. We emphasize that this secondary search (which will be referred to as \emph{\textbf{neighborhood search}} from now on) occurs on the existing graph, $G_N$, during which $G_N$ remains unchanged.
A pseudo-code for the neighborhood search (\textit{computePNS} routine) is provided in Algorithm~\ref{alg:computeNeighborhood}.

\begin{figure}
    \centering
    \includegraphics[width=0.5\linewidth]{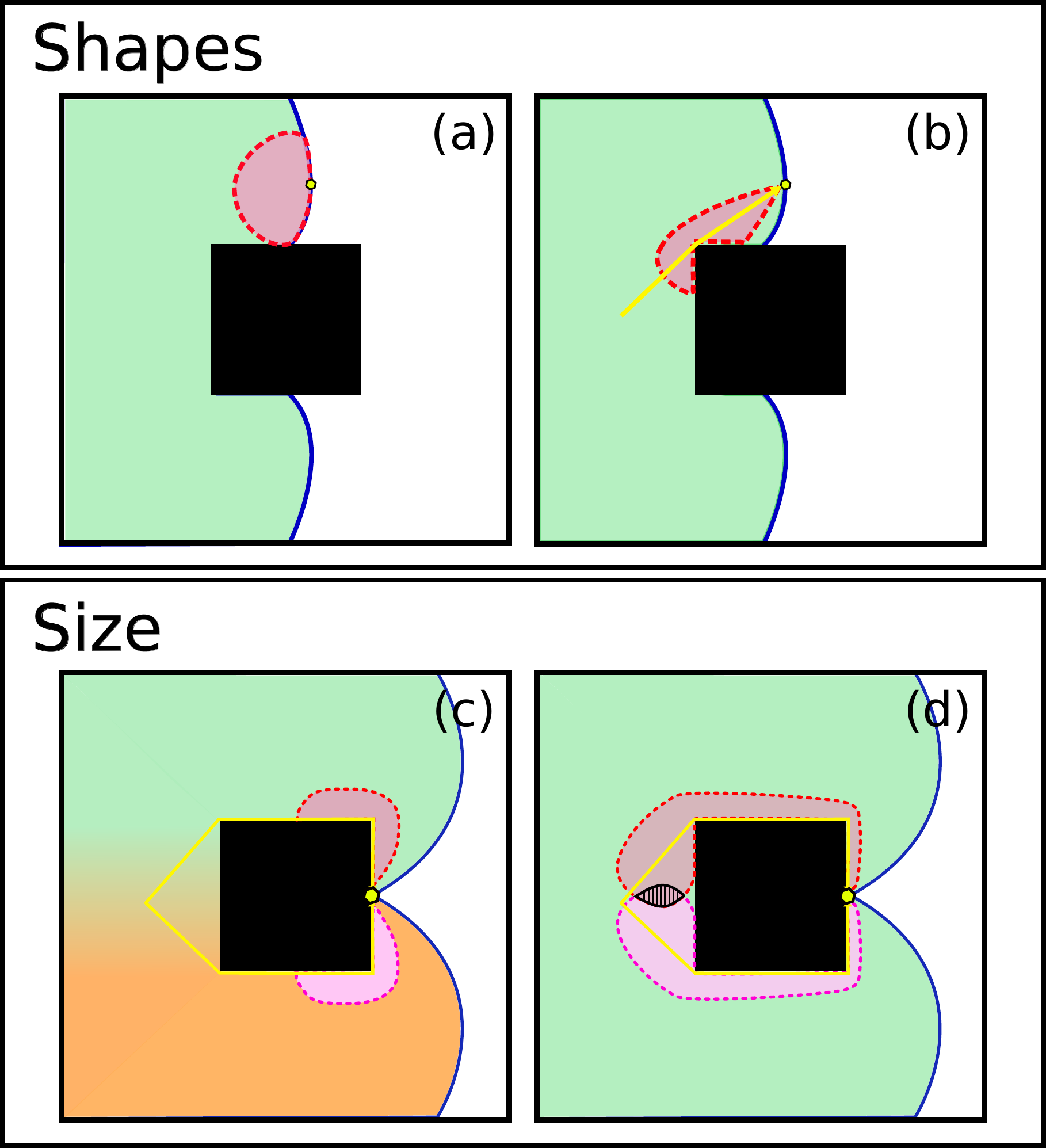}
    \caption{(a) Disk-shaped neighborhood at a vertex of interest (VOI). (b) \textit{Path-hugging} neighborhood at a VOI (Default option for path neighborhood sets). (c) Disjoint neighborhoods for two vertices at the same coordinates due to the obstacle resulting in the second copy to be inserted to the neighborhood augmented graph. (d) Intersecting neighborhoods at the upstream of the same obstacle, when the neighborhood size is increased. The second vertex is equivalent to the first one, thus it will be discarded.}
    \label{fig:neighborhood_shape_size}
\end{figure}

If the neighborhood search is carried out until a distance $r_n$ from $v$ (reaching a g-score of $r_n$), the expanded vertices will form a shape similar to a disk-section around $v$, bounded by a circle of radius $r_n$ at the upstream and by the open list of the primary search at the downstream (Figure~\ref{fig:neighborhood_shape_size}(a)). However, a disk-shaped set does not provide a reasonable representation for a neighborhood around the path leading up to $v$. It is possible to modify the shape of the computed path neighborhood set by \changedSB{utilizing the existing $g$-score of vertices from the primary search}. By using the $g$-score from the primary search as heuristic function to guide the secondary search, we obtain a path neighborhood set that \emph{hugs} the path more closely.
\changedSB{In practice, we use a parameter, $\omega$ (called a \emph{heuristic weight}), to determine the influence of the said heuristic function (Line~\ref{ln:heuristic} of Algorithm~\ref{alg:computeNeighborhood}) -- a heuristic weight of $0$ will create a disk-shaped neighborhood, while positive and larger heuristic weights will create more path-hugging neighborhood sets. We choose a $\omega\in[0,1]$.}
Figure~\ref{fig:neighborhood_shape_size}(a,b) \changedSB{illustrates a comparison} between a disc-shaped neighborhood and a path-hugging neighborhood.

\input{pseudocodes/pseudo_computeNeighborhood.tex}

\begin{figure}
    \centering
    \includegraphics[width=0.5\linewidth]{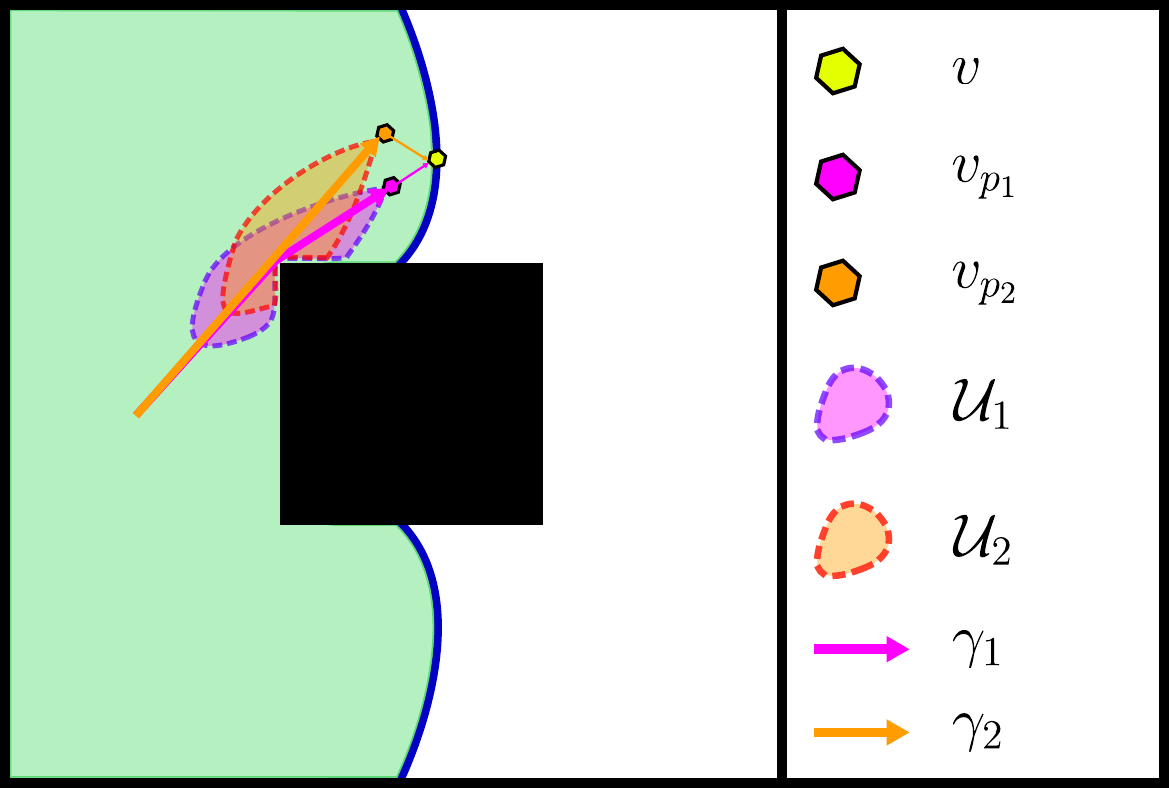}
    \caption{\changedASnew{A vertex at $v.q$ is generated by two distinct parent vertices $v_{p1}$ and $v_{p2}$. Shortest paths $\gamma_1$ and $\gamma_2$ leading up to the parent vertices are also distinct. However, the path neighborhood sets $U_1$ and $U_2$ intersect as the paths are only slightly different. In this case, there is no need to construct and maintain another vertex at the coordinates $v.q$.}}
    \label{fig:neighborhood_overlap}
\end{figure}

A \changedSB{path neighborhood set} of a vertex needs to satisfy following conditions to achieve desired branching behavior around the artifacts of the search space: 
\begin{enumerate}
    \item Vertices at the same coordinates generated through parent vertices that are close to each other in the graph should have large overlap \changedSB{in path neighborhood set}. Essentially, a branch of the search should not be giving rise to any subbranches unless it encounters an obstacle or a high-cost region. Size of the \changedSB{path neighborhood sets of nearby vertices in the same branch of a search} should be large enough to contain common vertices. \changedASnew{An example of this scenario is illustrated in Figure~\ref{fig:neighborhood_overlap}, where a vertex $v$, is generated by two different parent vertices $v_{p1}$ and $v_{p2}$ on the same branch whose path neighborhood sets have a large intersection. As a result, only a single vertex will be \changedSB{created and maintained at that point without creating any spurious branches}.}
    \item For an obstacle or a high-cost region to \changedSB{generate multiple topo-geometric branches during} the search, the size of \changedSB{path neighborhood set} should \changedSB{be smaller than the size of the obstacle/high-cost region} to ensure that there will not be any intersection at the upstream of the artifact (Figure~\ref{fig:neighborhood_shape_size}(c,d)). Hence, vertices generated via parent vertices on a different branch will be identified as different.
\end{enumerate}

The effect of the size of \changedSB{path neighborhood sets} relative to the obstacle size can be observed in Figure~\ref{fig:neighborhood_shape_size}(c,d). As observed empirically, disk-shaped \changedSB{path neighborhood sets} are more sensitive to the size parameter than the \textit{path-hugging} \changedSB{path neighborhood sets} as they tend to overlap at the upstream of an artifact even for smaller radii. Therefore, \textit{path-hugging} \changedSB{path neighborhood sets} are to be considered as the default \changedSB{path neighborhood sets} shape unless stated otherwise. 

\subsection{\changedSB{Theoretical Analysis}}
\label{sec:theory}

\subsubsection{\changedSB{A Sufficient Condition for Creating Topo-geometrically Distinct Branches}}

\changedASnew{We formalize the conditions for 
\changedSB{being able to compute topo-geometrically distinct branches at a vertex $q$}.
\changedSB{Let}
$r_s \in \mathbb{R}^+$ be the $g$-score at the open list at an instant of the search (under the assumption that vertices \changedSB{on the open list have same/similar} $g$-scores at the same instant).
\changedSB{We consider the graph to be a metric space with metric/distance function, $d$, so that for any $p_1,p_2\in V$, $d(p_1,p_2)$ is the total distance/cost of the shortest path through the set of already-expanded vertices connecting $p_1$ and $p_2$.}
\changedSB{Suppose a vertex $q\in V$ is being generated via two distinct geodsics/shortest paths (\emph{i.e.}, being reached from two different parents), $\gamma_1$ and $\gamma_2$ (see Figure~\ref{fig:proof}).

\begin{figure}
    \centering
    		\includegraphics[width=0.5\linewidth]{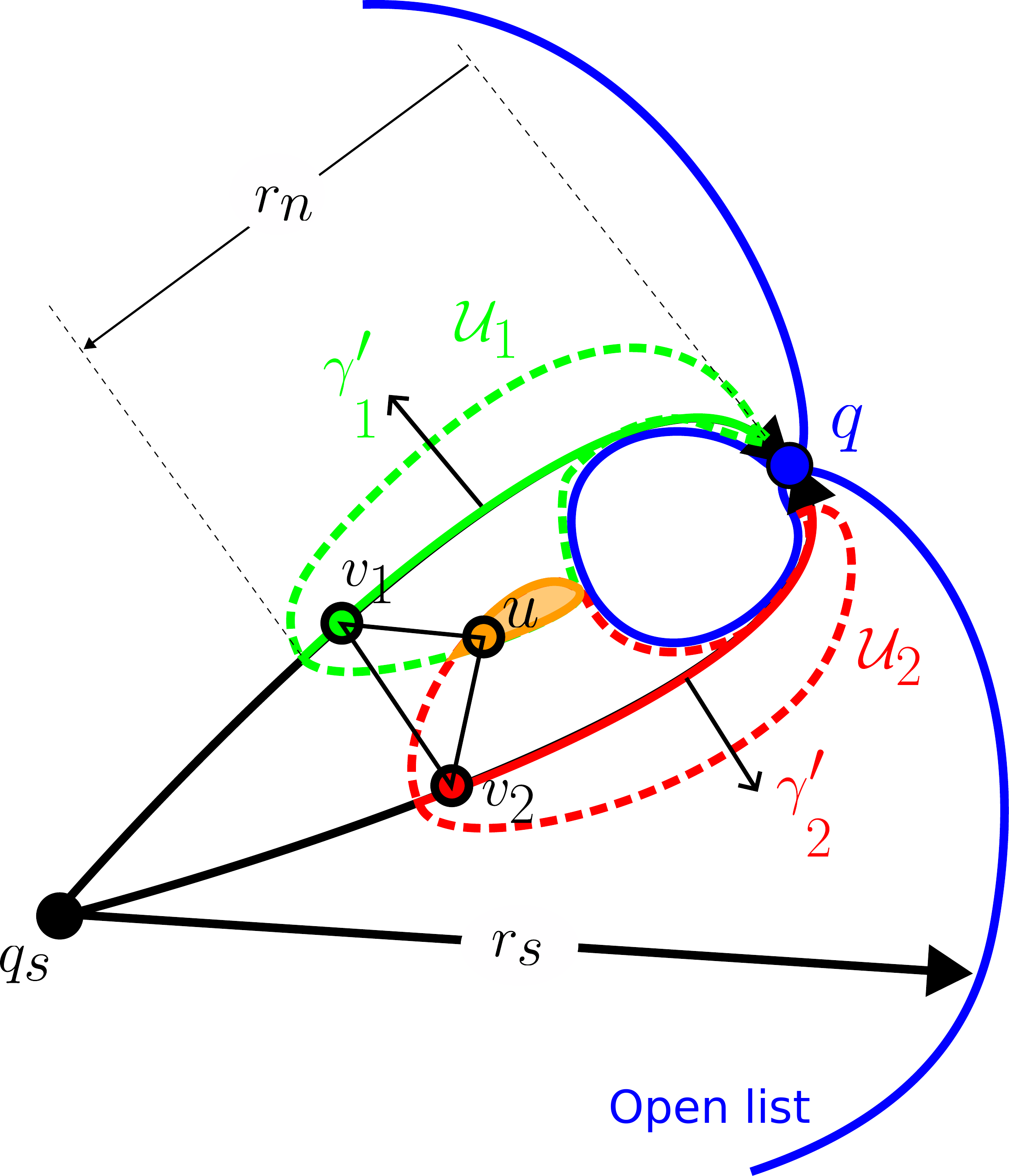} 
    	\caption{Path neighborhood sets $\mathcal{U}_1$ and $\mathcal{U}_2$ are shown for two paths $\gamma_1,\gamma_2$ for a point $q$ in the open list. Neighborhood radius is $r_n$ and the radius at the open list is $r_s$. $v_1$ and $v_2$ are points on attached path segments $\gamma_1'$ and $\gamma_2'$ . $u$ is a common neighbor, such that $u \in \mathcal{U}_1 \cap \mathcal{U}_2$.}\label{fig:proof}
    \end{figure}

We try to understand the condition under which the path neighborhoods of these two different paths, $\mathcal{U}_1$ and $\mathcal{U}_2$, will not overlap, hence giving rise to two distinct vertices $(q,\mathcal{U}_1)$ and $(q,\mathcal{U}_2)$, in the neighborhood augmented graph.
Let $\gamma_i'$ be the segment of the path, $\gamma_i$, contained within $\mathcal{U}_i$, $i=1,2$. Then the following result holds:

\begin{proposition}\label{proposition}
    ~
    \begin{center}
    If ~$\text{sep}(\gamma_1',\gamma_2') \geq \frac{4 r_n}{1-\omega}$, ~then~
    $\mathcal{U}_1 \cap \mathcal{U}_2 = \emptyset$.
    \end{center}
    where, $\text{sep}(\gamma_1',\gamma_2') = \inf_{v_1 \in \gamma_1', v_2 \in \gamma_2'} d(v_1,v_2) $ is the separation between the path segments $\gamma_1'$ and $\gamma_2'$.
\end{proposition}

The proof of this proposition is deferred to the Appendix. The proposition, in effect, gives a value for the minimum separation between two equal-length geodesic paths leading to the same vertex that will result in identification of the vertices $(q,\mathcal{U}_1)$ and $(q,\mathcal{U}_2)$ to be topo-geometrically distinct, and hence will give rise to two distinct branches in the neighborhood-augmented graph.

\begin{figure*}
    \centering
    \includegraphics[width=\linewidth]{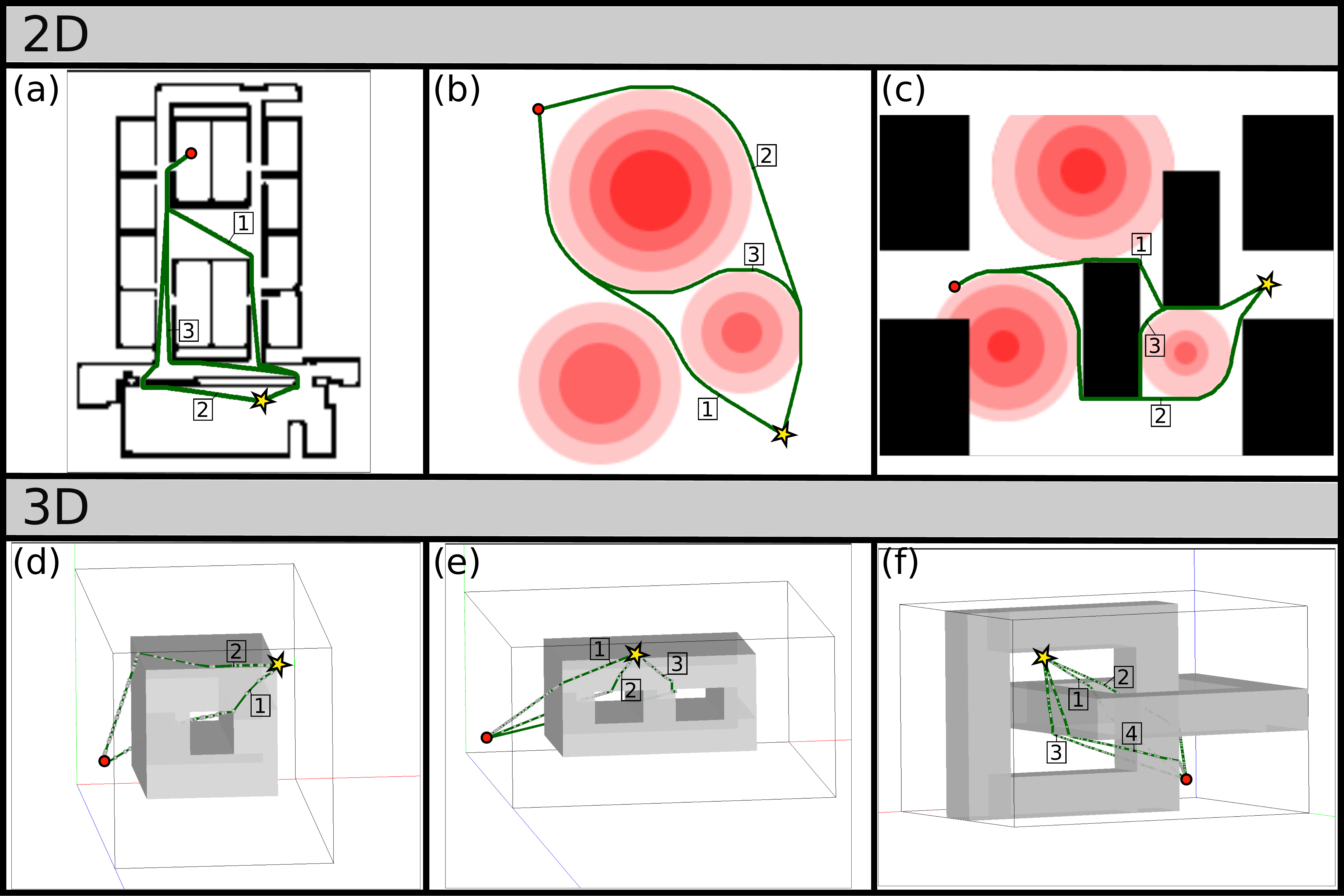}
    \caption{\changedASnew{\textbf{Results:} (a) Distinct paths found in a $(125\times146)$ 2D building-like environment (3 paths). Path lengths: $[141.57,142.72,154.72]$. Found in $7$ seconds. (b) $(150\times200)$ Environment with multiple hills (high-cost centers) (3 paths). Path lengths: $[217.93,273.01,280.87]$. Found in $35$ seconds. (c) $(200\times150)$ Environment with a mixture of obstacles and hills (3 paths). Path lengths: $[168.48,213.05,231.71]$. Found in $20$ seconds. (d) $(10\times10\times20)$ 3D environment with genus-1 object (2 paths). Path lengths: $[20.74,21.51]$. Found in $5$ seconds. (e) $(20\times10\times20)$ 3D environment with genus-2 object (3 paths). Path lengths: $[20.61,21.30,26.23]$. Found in $17$ seconds. (f) $(28\times20\times20)$ 3D environment with a chain-like structure (4 paths). Path lengths: $[30.97,32.20,32.44,34.53]$. Found in $60$ seconds.}}
    \label{fig:neighborhood_results}
\end{figure*}




}

}

\subsubsection{\changedSB{Complexity of the Neighborhood Augmented Graph Search}}

\changedSB{The Algorithm~\ref{alg:NASearch} is essentially a search in the neighborhood-augmented graph (in particular, it's illustrated with the Dijkstra's search algorithm, although any graph search algorithm can be used). The main difference is the computation of the path neighborhood set every time a vertex in the neighborhood-augmented graph is generated. The construction of the path neighborhood set of each vertex involves running a separate A* search up to a fixed distance of $r_n$ in the partially-constructed neighborhood-augmented graph, and hence by itself is a constant-time process on an average (while the path neighborhood sets of different vertices may contain different number of vertices depending on the cost in the neighborhood, in an uniform-cost region this number is almost constant, and on a high cost region this number is lower). Thus, the complexity of the search in the neighborhood-augmented graph is same as the complexity of the search algorithm used, only with a constant multiple. For example, the complexity of the Dijkstra's search in a graph with an almost-uniform vertex degree (e.g., graph constructed from uniform discretization of a configuration space and by connecting each vertex with its neighbors) is $O(|\mathcal{V}| \log(|\mathcal{V}|))$, where $|\mathcal{V}|$ is the number of vertices expanded. The complexity of searching in the neighborhood-augmented graph using Dijkstra's search is thus also $O(k_n |\mathcal{V}| \log(|\mathcal{V}|))$, where the constant $k_n$ is the constant-time complexity of generating the path neighborhood of each vertex.}

\subsection{Implementation Details}\label{subsec:implementation_details}

The main ideas and procedures of the search on a neighborhood augmented graph is as described in previous sections and the pseudocodes within. \changedASnew{In this section, we will \changedSB{describe} some implementational details that are not captured in previous sections.}

\textbf{Rollback to grandparent.} \changedAS{In previous sections, it is assumed that the neighborhood search for a vertex $v'$ starts from its parent vertex $v$. As a result, \changedSB{path neighborhood sets} are always illustrated as attached to the vertex. However, depending on the primary search algorithm being used, at the time of neighborhood search, vertex $v$ might not be inserted in the neighborhood-augmented graph, thus making it impossible to perform a search starting from $v$ (this mainly occurs when using S* search algorithm, which generates the grandchildren of a vertex when expanding a vertex). A solution is to \textit{rollback} from the vertex $v'$ for some fixed number of generations (which means to consider the parent or $n$-th grandparent of the vertex) and start the neighborhood search from that vertex. This procedure is performed by the \textit{rollback} subroutine in Line~\ref{ln:rollback} of Algorithm~\ref{alg:computeNeighborhood}, which recursively calls the `came\_from' attribute of a vertex for desired number of times.}

\changedASnew{\textbf{Neighborhood search depth.} In previous sections, neighborhood radius $r_n$ was expressed in terms of the g-scores on the corresponding graph. However, in an environment that is uniformly discretized, on a high-cost region, a neighborhood whose radius is based upon the g-scores \changedSB{only may} contain very few vertices. 
To alleviate this situation, we place a lower bound on the depth \changedSB{(in terms of number of generations/hops from the start vertex of the neighborhood search) that must be performed for a the neighborhood search.}}



\subsection{Results} \label{sec:results-1}
\textbf{Results in 2D and 3D domains:}
\changedASnew{Proposed neighborhood augmented path planner is validated in 6 different planning environments shown in Figure~\ref{fig:neighborhood_results}. The algorithm is run on an 11th Gen Intel Core i7-1165G7 @ 2.80GHz with 16GB of RAM. The primary search is performed using S* and the neighborhood searches are performed using A* algorithms. 2D and 3D environments are discretized into uniform triangles and tethrahedrons respectively. On nonuniform cost environments, the cost of a line segment lying entirely in white space is equal to the euclidean norm, whereas a line segment lying entirely in the high cost center (marked with darkest shade of red) has a cost scaled by a factor $\sim3$. For both 2D and 3D environments, heuristic weight is set as $\omega = 0.6$. On 2D environments, we use the rollback radius $r_b=4$ and neighborhood radius $r_n=10$. On 3D environments, we use the rollback radius $r_b=3$ and neighborhood radius $r_n=5$.

As shown in Figure~\ref{fig:neighborhood_results}(a-f), neighborhood augmented path planning algorithm can identify locally optimal and topo-geometrically distinct paths in 2D environments with obstacles, high-cost regions, or a mixture of both. In 3D environments, it can identify distinct paths in the presence of objects with non-zero genus, or even more complex structures such as knots and chains. The local suboptimality of some of the paths obtained in 3D environments are mainly an artifact of the coarse discretization \changedSB{used with S* search algorithm}.}

\begin{figure}
    \centering
    \includegraphics[width=\linewidth]{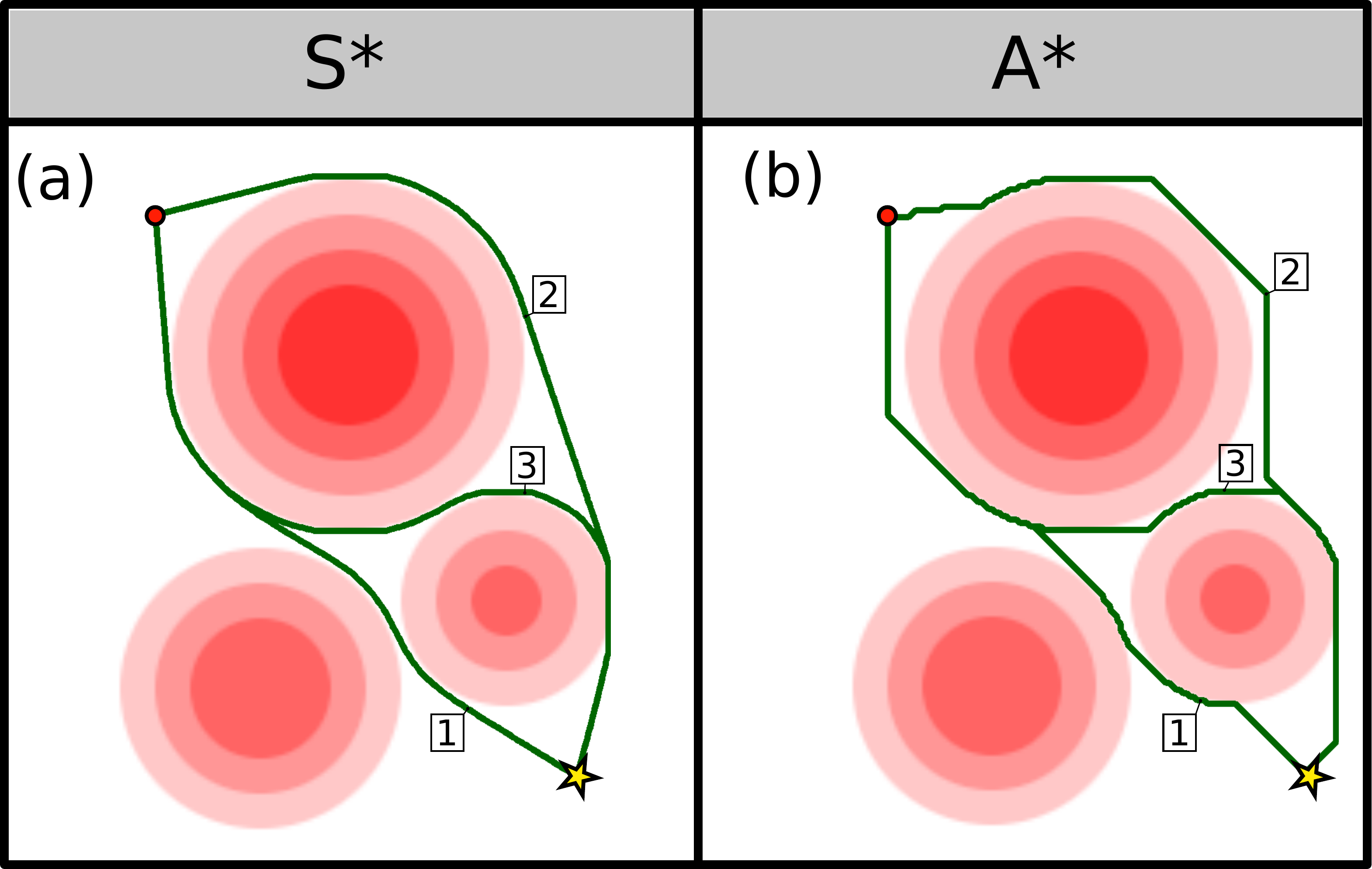}
    \caption{\changedASnew{\textbf{Results:} (a) Paths found using S*. Path lengths: $[217.93,273.01,280.87]$. Found in $35$ seconds. (b) Paths found using A*. Path lengths: $[223.77,285.83,288.18]$. Found in $49$ seconds.}}
    \label{fig:SStar_AStar_comparison}
\end{figure}

\changedSB{\textbf{Comparison between S* and A* as the primary search algorithm:}
The neighborhood-augmented graph can be searched with any search algorithm, including Dijkstra's, A* and S*. Figure~\ref{fig:SStar_AStar_comparison} shows the comparison of the results obtained in an 2D domain with non-uniform cost using the S* and the A* search algorithms. The A* search algorithm uses an 8-connected grid-world representation of the domain as the discrete graph, $G$, using which the neighborhood-augmented graph, $G_N$, is incrementally constructed for the search and computation of 3 topo-geometrically distinct paths.}

\begin{figure}
    \centering
    \includegraphics[width=0.75\linewidth]{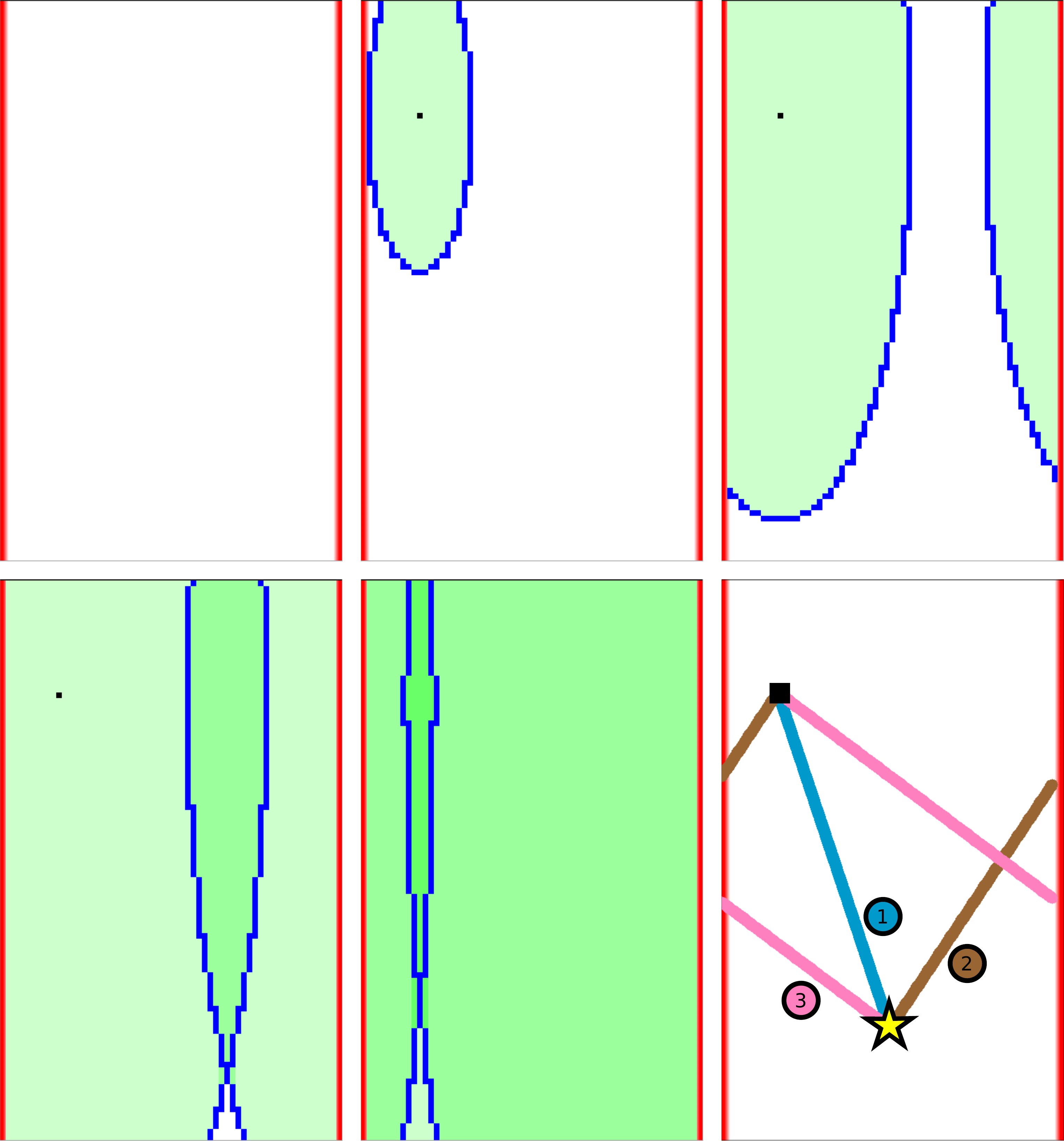}
    \caption{\textbf{Results:} \changedASnew{Path planning on a cylindrical surface, $\mathbb{S}\times\mathbb{R}$, with $r_c=30$ and $h=100$. The cylinder is represented as a rectangle, whose left and right edges are identified/connected (colored red). Neighborhood-augmented path planning algorithm is \changedSB{used to compute} 3 \changedSB{topo-geometrically distinct} paths.
    Path lengths from search: [86.87, 139.25, 258.39].
    These path lengths match the analytically computed path lengths.
    }}
    \label{fig:cylinder_planning}
\end{figure}

\changedASnew{
\textbf{Path planning on spaces with non-Euclidean topology:} 
We \changedSB{also} demonstrate the capabilities of the neighborhood augmented path planning algorithm in a
\changedSB{space that is not a subset of a Euclidean space.} 
\changedSB{We choose the surface of a cylinder, $\mathbb{S}\times\mathbb{R}$, to demonstrate this capability}, with radius $r_c$ and height $h$. The surface is represented by a $2\pi r_c \times h$ rectangle, with \changedSB{the two} vertical edges identified. The search procedure and the resulting paths are shown in Figure~\ref{fig:cylinder_planning}. \changedSB{We computed $3$ paths through search in the neighborhood-augmented graph,} although clearly it is capable of finding any number of paths on the cylinder simply by \changedSB{allowing the search to continue further and allowing the search front to wind} around the cylinder \changedSB{the} required number of times. Results show that the lengths of the paths obtained by the algorithm match with the path lengths computed analytically.}


\textbf{Cost multiplier analysis:} As mentioned, on uniform cost and flat subsets of Euclidean planes (for example, Figure~\ref{fig:neighborhood_results}(a)), the path cost between two vertices is computed using the Euclidean distance as $c(v_i,v_j) = d(v_i,v_j)$, where $d(v_i,v_j) = ||v_i-v_j||_2$. For environments with high-cost regions (as in Figure\ref{fig:neighborhood_results}(b,c)), the path cost between two vertices is implemented as a function of the color of the corresponding cells and a cost multiplier (CM). Let $\rho(v):V\to[0,1]$ map the given vertex to the color value. Then the path cost between two vertices in a nonuniform path cost environment is computed as $c(v_i,v_j) = (v_i,v_j)\left[1+\text{CM}\frac{\rho(v_i)+\rho(v_j)}{2}\right]$.

\begin{figure}
    \centering
    \includegraphics[width=0.75\linewidth]{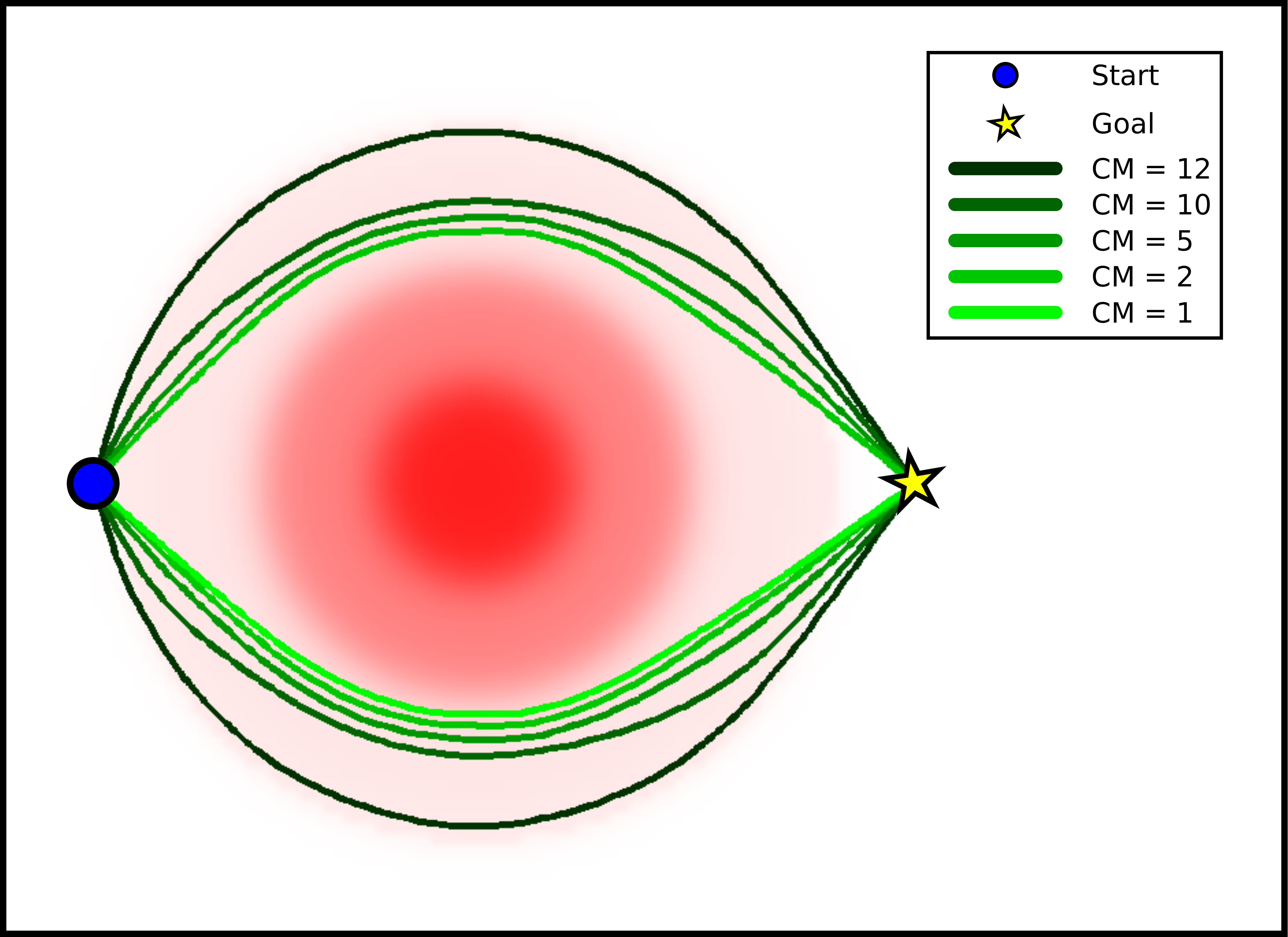}
    \caption{\textbf{Results:} Shortest paths obtained on a nonuniform cost environment. As cost multipliers (CM) decrease, shortest paths tend to get closer to the high cost region. For CM$=1$, the algorithm can only find a single path.}
    \label{fig:cm_analysis}
\end{figure}

The geometry of the optimal paths obtained by the neighborhood augmented planning method varies with the cost multiplier being used in the environment. Figure~\ref{fig:cm_analysis} shows how locally optimal paths tend to get closer to the center of the high cost region as the cost multiplier decreases. It is worth noting that above a certain value of the cost multiplier paths will not move further away from the high cost region. At the other extreme (if the cost multiplier is below a certain value), proposed algorithm will become incapable of distinguishing between multiple optimal paths and return only one of them (see the path for $\text{CM}=1$ in Figure~\ref{fig:cm_analysis}). This issue is explained further in detail and tackled in Section~\ref{sec:merge_points}.

%% file: pseudocodes/pseudo_NAsearch.tex
\begin{algorithm}
	\caption{Incremental Construction and Dijkstra's Search in a Neighborhood-augmented Graph (NAG)} \label{alg:NASearch}
        \begin{algorithmic}[1]
        \Statex $G_N =$ \textit{searchNAG}~($q_s,\mathcal{N}_G,\mathcal{C}_G, @\text{\textit{stopSearch}}$)
    \Statex \hrulefill
         \Statex Inputs:
         \begin{enumerate}[leftmargin=2em,label=\alph*.]
         \item Start configuration $q_s \in V$
         \item Neighbor/successor function $\mathcal{N}_G$ (this decsribes the connectivity of graph $G$)
        \item Cost function $\mathcal{C}_G: V \times V \to \mathbb{R}^+$
        \item Stopping criteria (function), \textit{stopSearch}: $V_N\rightarrow \{0,1\}$
        \end{enumerate}
    \Statex Output:
    \Statex \hspace{2em} Graph $G_N$, with costs and neighborhoods computed for every vertex
    \Statex \hrulefill
    \State $v_s := (q_s,\{\& v_s\})$ \algcomment{start vertex in $V_N$, with self-reference} 
    \Statex \phantom{$v_s := (q_s,\{\& v_s\})$} \algcomment{in its path neighborhood set.}
    \State Set $g(v_s) = 0$ \algcomment{g-score}
    \State $V_N = \{v_s\}$ \algcomment{vertex set}
    \State $E_N = \emptyset$ \algcomment{edge set, maintained implicitly as a link tree/graph}
    \State $Q = \{v_s\}$ \algcomment{open list, maintained by a heap data structure.}
    \State $v:= v_s$
    \While {$Q \neq \emptyset$ \textbf{AND} \textbf{not} \textit{stopSearch}($v$)}
        \State $v := (q,\mathcal{U}) = \arg\!\min_{v' \in Q} \, g(v')$ \algcomment{heap pop.}
        \State $Q = Q - v$ \algcomment{heap pop.}
        \State $\mathcal{U}' =$ \textit{computePNS}\,($v, G_N$) \algcomment{path neighborhood set} \label{ln:computePNS} 
        \Statex \phantom{$\mathcal{U}' =$ \textit{computePNS}($G_N,v$)} ~~~~\algcomment{for path leading to $v$}
        \For {$q' \in \mathcal{N}_G(v)$}
            \State $v' := (q',\mathcal{U}')$ \algcomment{potential neighbor/successor}
            \State $g' = g(v) + \mathcal{C}_G(q,q')$ \algcomment{potential g-score for $v'$}
            \If {$\nexists~ w \in V_N$, with $v' \equiv w$} \algcomment{new vertex}
                \State $V_N = V_N \cup \{v'\}$
                \State $E_N = E_N \cup \{(v,v')\}$ \algcomment{maintained as linktree}
                \Statex \phantom{.}\qquad\qquad \algcomment{for neighbor function computation, $\mathcal{N}_{G_N}$}
                \State $g(v') = g'$
                \State $Q = Q \cup \{v'\}$
                \State $v'.\text{came\_from} = v$
            \Else \algcomment{vertex already exists ($w$)} \label{ln:elseMP}
                \State $E_N = E_N \cup \{(v,w)\}$
                \If {$g' < g(w)$ \textbf{AND} $w\in Q$} \algcomment{update $w$}
                    \State $g(w) = g'$
                    \State $w.\text{came\_from} = v$
                    \State $w.\mathcal{U} = \mathcal{U}'$
                \EndIf
            \EndIf
            \EndFor
    \EndWhile
    \State $G_N := (V_N,E_N)$
    \State \textbf{return} $G_N$
	\end{algorithmic} 
\end{algorithm}

%% file: pseudocodes/pseudo_stopSearchGoal.tex
\begin{algorithm}
	\caption{Goal-Based Stopping Criteria for Search} \label{alg:stopSearchGoal}
        \begin{algorithmic}[1]
        \Statex [bool, $\mathcal{V}_g$] $=$ \textit{stopSearch\_AtGoal}${}_{q_g,n_p}$($v$)
    \Statex \hrulefill
         \Statex Input: Current vertex $v = (q,\mathcal{U})\in V_N$
        \Statex Static variables / parameters:  \begin{enumerate}[leftmargin=2em,label=\alph*.]
        \item Goal configuration $q_g \in V$
        \item Number of paths to find $n_{p}$
        \end{enumerate}
    \Statex Output:
    \begin{enumerate}[leftmargin=2em,label=\alph*.]
            \item Boolean (\textbf{true} to stop search, \textbf{false} to continue)
            \item Vertices in the NAG found at goal configuration $\mathcal{V}_{g}$
            \end{enumerate}
    \Statex \hrulefill
    \State \textbf{static} $\mathcal{V}_{g} (= \emptyset)$ \algcomment{static variable initiated with empty set}
    \If {$v.q = q_g$ \textbf{AND} $v \notin \mathcal{V}_{g}$}
        \State $\mathcal{V}_{g} = \mathcal{V}_{g} \cup {v}$
        \State \textbf{return} ~$(|\mathcal{V}_{g}| == n_{p})$
    \EndIf
	\end{algorithmic} 
\end{algorithm}

%% file: pseudocodes/pseudo_computeNeighborhood.tex
\begin{algorithm}
	\caption{Path Neighborhood Set (PNS) Computation using A* Search} \label{alg:computeNeighborhood}
        \begin{algorithmic}[1]
        \Statex $\mathcal{U} =$ \textit{computePNS}${}_{\{r_n, \omega, r_b\}}$($v_p, G_N$)
    \Statex \hrulefill
         \Statex Inputs:
         \begin{enumerate}[leftmargin=2em,label=\alph*.]
         \item Existing/current neighborhood-augmented graph $G_N=(V_N,E_N)$, along with the g-scores of vertices in the graph, $g(\cdot)$.
         \item Parent vertex $v_p = (q_p,\mathcal{U}_p) \in V_N$
        \end{enumerate}
         \Statex Geometry Parameters:
         \begin{enumerate}[leftmargin=2em,label=\alph*.]
         \item Neighborhood radius, $r_n \in \mathbb{R}^+$
         \item Heuristic weight, $\omega \in [0,1]$
          \item Rollback amount, $r_b\in \mathbb{Z}_{\geq0}$
        \end{enumerate}
    \Statex Output:
    \Statex \hspace{2em} A set of vertices (\textit{path neighborhood set}), $\mathcal{U} \subset V_N$
    \Statex \hrulefill
    \State $v_s := \textit{rollback}(v_p,r_b)$ \algcomment{start vertex for secondary search}  \label{ln:rollback}
    \State $\mathcal{U} := \emptyset$
    \State $\widetilde{g}(v) = \infty, \widetilde{f}(v) = \infty, ~\forall v\in V_N$ \algcomment{implicitly set all f- and g- score for secondary search to infinity for all vertices}
    \State $\widetilde{g}(v_s) := 0$ \algcomment{g-score in secondary search}
    \State $\widetilde{f}(v_s) := \widetilde{g}(v_s) + \omega \,g(v_s)$, 
    \algcomment{f-score in secondary search. $g$ refers to the g-score from the primary search, which is being used as a heuristic function for the secondary search}
    \State $\mathcal{U} = \{v_s\}$ \algcomment{vertex set for neighborhood search}
    \State $\widetilde{Q} := \{v_s\}$ \algcomment{open list for neighborhood search}
    \State $v:= v_s$
    \While {$Q \neq \emptyset$ \textbf{AND} \textbf{not} 
    $\widetilde{g}(v) > r_n$
    }
        \State $v := \arg\!\min_{v' \in \widetilde{Q}} \widetilde{f}(v')$
        \State $\widetilde{Q} = \widetilde{Q} - v$
        \For {$v' \in \mathcal{N}_{G_N}(v)$} 
            \State $\bar{g}' = \widetilde{g}(v) + \mathcal{C}_{G_N}(v,v')$
            \If {$\bar{g}' < \widetilde{g}(v')$} \algcomment{better g-score}
                \State $\widetilde{g}(v') = \bar{g}'$
                \State $\widetilde{f}(v') = \bar{g}' + \omega \,g(v')$ \label{ln:heuristic}
                \State $\widetilde{Q} = \widetilde{Q} \cup \{v'\}$
                \State $\mathcal{U} = \mathcal{U} \cup \{v'\}$
            \EndIf
        \EndFor
    \EndWhile
    \State \textbf{return} $\mathcal{U}$
	\end{algorithmic} 
\end{algorithm}

%% file: sections/merge_points_TBD.tex
\subsection{\changedSB{Failure of Neighborhood-augmented Search in Low Cost/Curvature Domains}}
\begin{figure}
    \centering
    \includegraphics[width=0.3\textwidth]{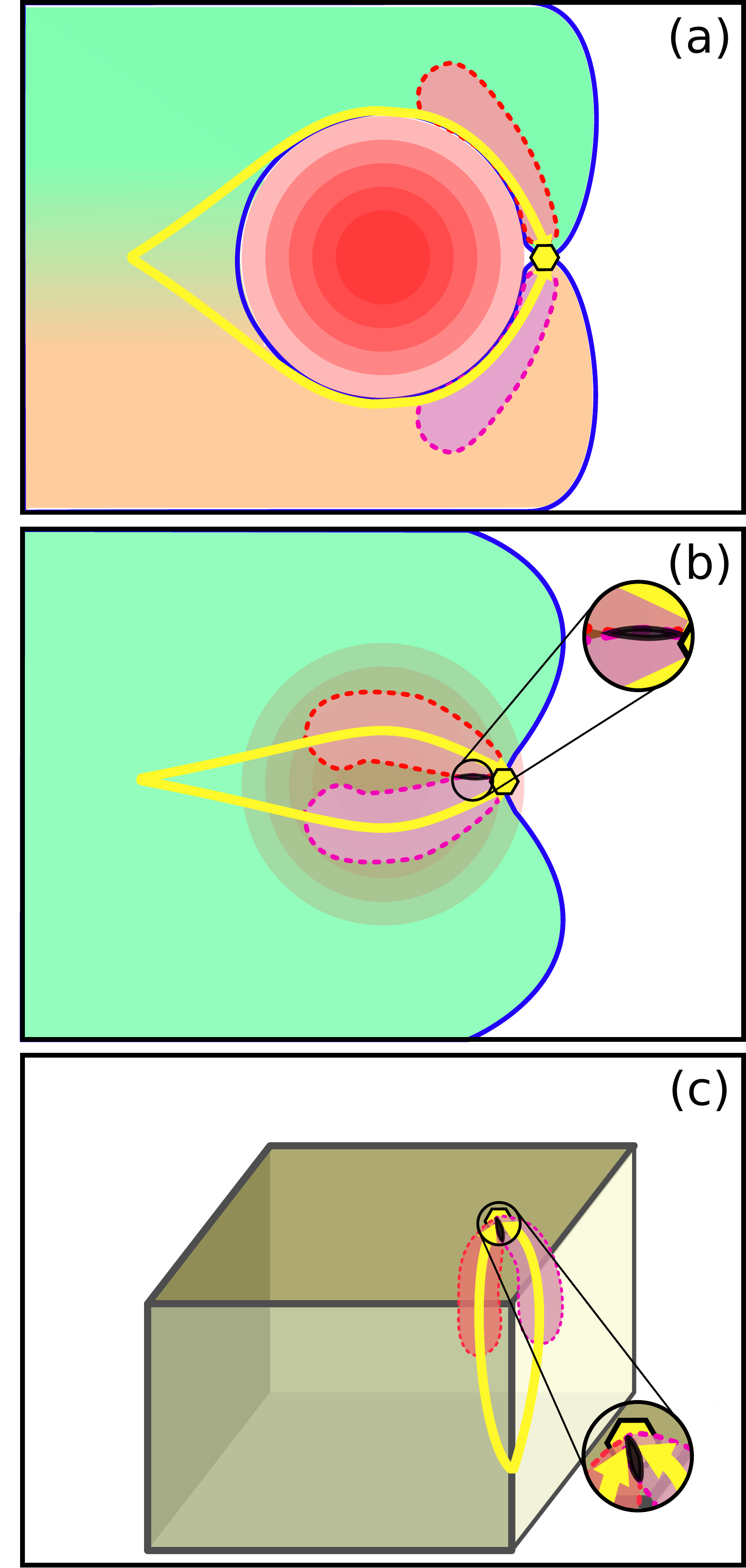}
    \caption{(a) Through a steep hill (high-cost center with high value) search propagation is significantly slowed down, thus two branches emerge. Neighborhoods remain separate for vertices reached via different branches. (b) Around a flatter hill, (high-cost center with smaller value) neighborhoods maintain an overlap, hence branches do not emerge naturally even on a neighborhood augmented graph (neighborhoods maintain the overlap, because the flatter hill only slows down the search through its center by a small amount, causing a small cusp in the search wave). (c) Same/similar behavior is observed around a corner of a 3D prism.}
    \label{fig:neighborhood_limitations}
\end{figure}

As explained in Section~\ref{subsec:neighborhood_augmented_graph}, comparison of neighborhoods allow distinction between vertices with the same coordinates that are reached via different branches of the search. However, the branching of the search is only possible in the presence of obstacles and high-cost regions in 2D, and elongated or higher-genus obstacles in 3D. As the curvature of an environment 
decreases (\changedSB{e.g., for a high-cost region in a planar domain if the cost is not sufficiently high}), the path neighborhoods keep overlapping and is unable to give rise to distinct topo-geometric branches (this is primarily due to the fact that the path neighborhoods are of finite size and width, which is necessary in the first place to check overlaps).
Instead \changedSB{of creating distinct branches}, only a small cusp in the search wave-front is generated, \changedSB{and the search in the neighborhood-augmented graph is unable to compute multiple topo-geometrically distinct paths (Figure~\ref{fig:cm_analysis}, with CM=1)}. Similarly, around a corner of a prism in 3D, the search is only slowed down by a small amount through the corner. 
This behavior is illustrated in Figure~\ref{fig:neighborhood_limitations} (b,c). In these cases, only a single path is \changedSB{found} as a result of the search on the neighborhood augmented graph.

\subsection{\changedSB{Cut} Points}

\changedASnew{On a flat \changedSB{planar domain} without any holes there is a unique geodesic between any two points and it is neither necessary nor expected to generate different branches of the search. As some small curvature is introduced to the surface \changedSB{by a high-cost region}, multiple locally optimal paths \changedSB{can} arise between two points. However, along the cusp generated in the search wave by the low curvature, these paths are very similar to each other and the corresponding path neighborhood sets are likely to intersect. That is why it is not possible to generate branches of the search in low curvature environments \changedSB{(Figure~\ref{fig:neighborhood_limitations}(b))}.

Although it is possible to modify the shape and size of the path neighborhood sets, it is not possible to come up with a neighborhood radius and heuristic weight that allow distinction between such close paths without causing spurious topo-geometric branches to be generated at other locations in the environment. A similar behavior is observed around the corners of a 3D prism. Another approach could be to consider a finer discretization, to increase the capability of distinguishing between similar paths. However, we propose an approach to tackle with the low curvature environments, without requiring any changes to discretization or further fine tuning of the neighborhood parameters.

The idea behind the proposed solution is to introduce an artificial \emph{cut} to the \changedSB{neighborhood-augmented graph} that will allow separate branches of the search to be generated. To decide where the cut should be introduced, we identify the points along the cusp generated by the low curvature to which the distinct geodesics are separated significant enough. These points will be referred to as \emph{\changedSB{cut} points}. Then, it would be possible to treat a region around those points (referred to as \emph{\changedSB{cut} point regions} (CPR)) as an obstacle to give rise to different branches of the search. 

\begin{figure}
    \centering
    \includegraphics[width=0.3\textwidth]{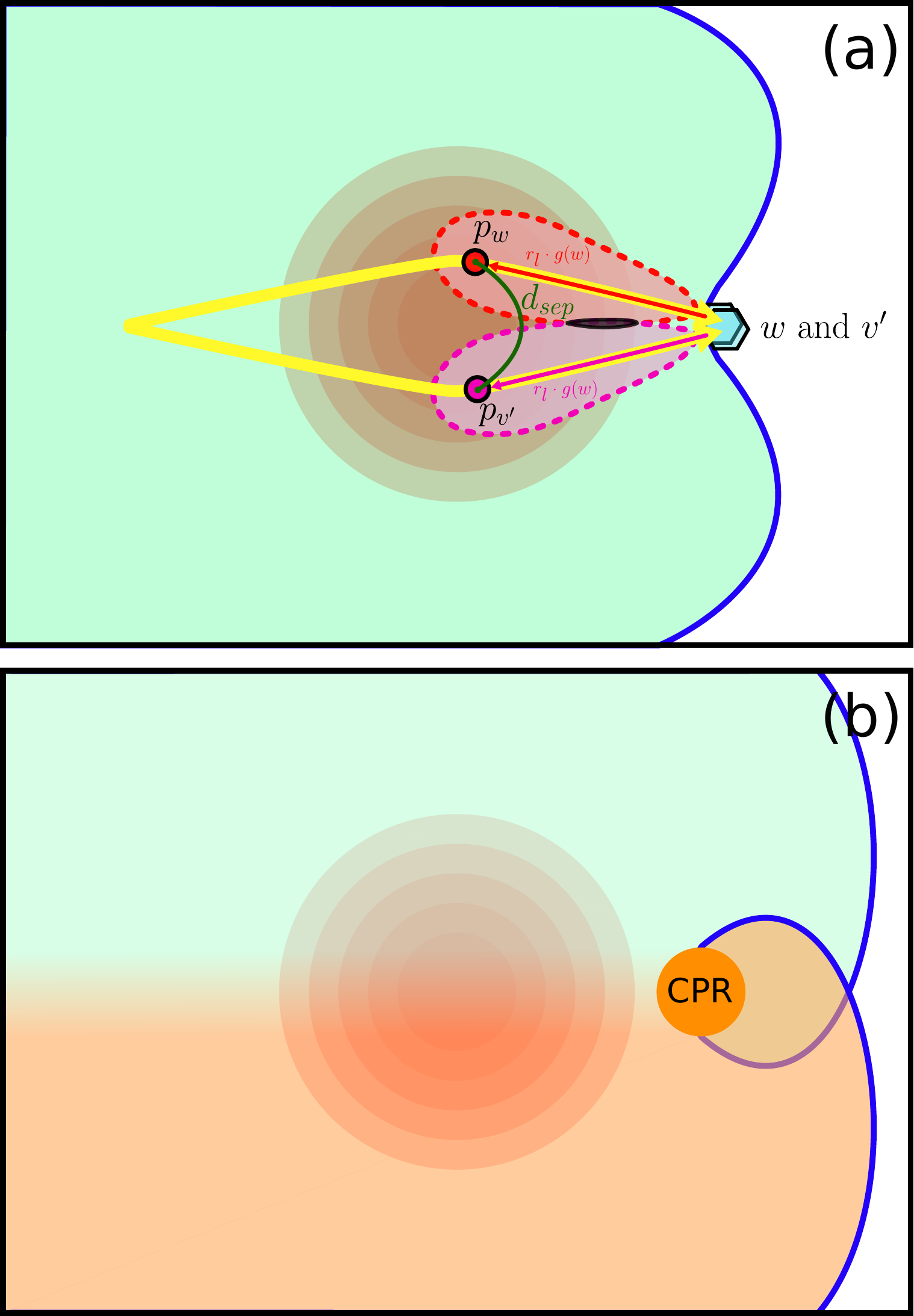}
    \caption{(a) Visualization of two vertices $w$ and $v'$ with small neighborhood intersection. Shortest paths are constructed until a distance of $r_l\cdot g(w)$. Resulting vertices $p_w$ and $p_{v'}$ have a separation of $d_{sep}$ in between. (b) On a low curvature environment, \changedSB{cut} point region is detected at the downstream of the hill. Treatment of these \changedSB{cut} points causes branches to emerge around this region.}
    \label{fig:merge_points}
\end{figure}

A pseudocode describing the \emph{\changedSB{cut} point} identification and \changedSB{cut} point region generation procedures are given in Algorithm~\ref{alg:mergePointCheck}. When running the search algorithm on low curvature environments, the \textit{\changedSB{cutPointCheck}} subroutine is called within the \textit{searchNAG} routine provided in Algorithm~\ref{alg:NASearch}, after a successor $v'$ is identified the same with an existing vertex $w$ -- after Line~\ref{ln:elseMP}. To identify the \changedSB{cut} points algorithmically, we utilize the following observation. Although the path neighborhood sets for vertices at the cusp are not entirely disjoint, they should have relatively smaller intersections. It is possible to compute a neighborhood intersection ratio (ratio of the number of common vertices in the path neighborhood sets to the size of the path neighborhood sets), such that the ratios at a \changedSB{cut} point falls below a certain threshold -- Line~\ref{ln:getNeighborhoodIntersectionRatio} of Algorithm~\ref{alg:mergePointCheck}. Parallel to that, we also make use of the observation that the lengths of the geodesics to two vertices should be similar to each other at the cusp (this can be confirmed by looking at the difference in g-scores of the vertices) -- Line~\ref{ln:gscorecheck} of Algorithm~\ref{alg:mergePointCheck}. Then we reconstruct a segment of the geodesics from vertices $w$ and $v'$ to the start vertex of the search $v_s$. Assuming that the length of the entire geodesic is equal to the g-score at the vertex (e.g. $g(w)$ for $w$), the end points of these segments are the path points $p_w$ and $p_{v'}$ which are at a distance of $r_l\cdot g(w)$ away from $w$ and $v'$ respectively, along the corresponding geodesics -- obtained at Lines~\ref{ln:ppw} and \ref{ln:ppv'} and illustrated in Figure~\ref{fig:merge_points}(a). For computational efficiency, we compare $p_w$ and $p_{v'}$ instead of reconstructing the entire paths and computing the max-min distances. To compare, we run a separate A* search starting from $p_w$ for a radius $\varepsilon_{upper}$ -- Line~\ref{ln:dsep}. If this search does not reach $p_{v'}$, this is interpreted as a large separation (caused by an obstacle or high-curvature). If the search reaches $p_w$, it is checked whether the length of the path is larger than some lower bound $\varepsilon_{lower}$, which ensures that identical vertices are not assigned as \changedSB{cut} points at spurious locations on the map -- Line~\ref{ln:boundcheckMP}. After detecting a \changedSB{cut} point, we perform another Dijkstra’s search for a fixed radius $r_{MP}$ and generate the \changedSB{cut} point region (CPR) consisting of the vertices lying within. The \changedSB{cut} point region is treated as an obstacle, leading to separate branches to be generated downstream as illustrated in Figure~\ref{fig:merge_points}(b).}


\input{pseudocodes/pseudo_mergePointCheck.tex}

\changedASnew{Essentially, the cut point check procedure allows us to make a distinction between geodesics that are too close to each other (which cannot be distinguished using path neighborhood sets without significant modifications to the neighborhood parameters or to the discretization). Therefore, one might argue that the same procedure could replace the neighborhood search and comparison. However, even for a segment of the path, the path reconstruction is computationally expensive \changedSB{(grows linearly with the length of the path or the distance from the start instead of being constant-time as is the cse for path neighborhoods) and computing distance between the path segments require additional searches}. Considering that neighborhood search and comparison procedures are being performed for every vertex, it would be infeasible to replace it with a procedure that explicitly reconstructs geodesics and computes the distance between them. \changedSB{Furthermore, path neighborhood overlap computation is less sensitive to small changes in the environment compared to computation of distance between shortest paths.}}

\begin{figure}
    \centering
    \includegraphics[width=\linewidth]{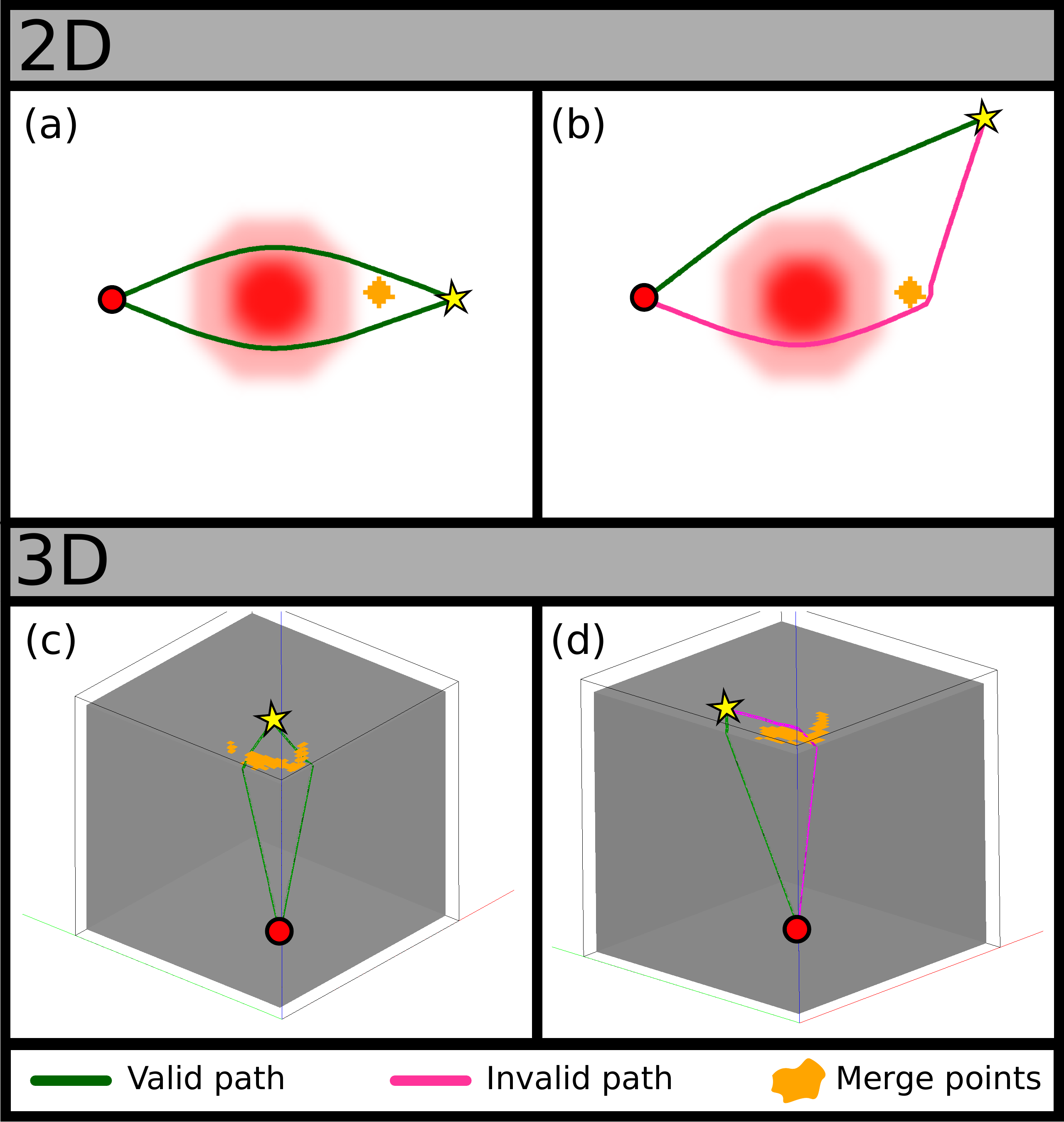}
    \caption{\textbf{Results:} (a) Two valid paths around a small \emph{hill} (high cost region, with low cost multiplier), when the start and goal are placed symmetrically. (b) A valid and an invalid path around a small hill, with start and goal placed asymmetrically. Invalid path touches the artificial cut. (c) Two valid paths around the corner of a prism, when the start and goal are placed symmetrically. (d) A valid an an invalid path around the corner, with start and goal placed asymmetrically. Invalid path goes around the artificial cut introduced by the \changedSB{cut} points.}
    \label{fig:merge_point_results}
\end{figure}

\subsection{\changedSB{Results with Cut Point Identification}}
\changedASnew{Path-planning using the neighborhood augmented graph with the addition of \changedSB{cut} point regions can identify distinct paths in 2D environments with smaller bumps. In a 3D environment, it can identify distinct path around a corner of a prismatic obstacle. Examples for 2D and 3D scenarios are provided in Figure~\ref{fig:merge_point_results}(a-d). For both 2D and 3D cases, heuristic weight $\omega=0.6$, rollback radius $r_b=3$, $r_l = 0.2$ and $\varepsilon_{g} = 0.1$ are used. For the 2D case, we use the neighborhood radius $r_n=8$, $\varepsilon_{lower} = 8$, $\varepsilon_{upper} = 25$, $\varepsilon_{i} = 0.6$, and $r_{MP} = 3$. For the 3D case, we use the neighborhood radius $r_n=6$, $\varepsilon_{lower} = 5$, $\varepsilon_{upper} = 10$, $\varepsilon_{i} = 0.4$, and $r_{MP} = 5$.

In both 2D and 3D cases, a high cost center or a prismatic corner gives rise to two shortest paths, when the start and goal are placed symmetrically with respect to the artifact. However, when placed asymmetrically, the second path can only be identified with the help of the \changedSB{cut} points (the artificial cut). Those paths that present abrupt cornering behavior around \changedSB{cut} point regions are considered as invalid as they are not locally optimal in the underlying configuration space.}



%% file: pseudocodes/pseudo_mergePointCheck.tex
\begin{algorithm}
	\caption{\changedSB{Cut} Point Check and \changedSB{Cut} Point Region Generation Procedure} \label{alg:mergePointCheck}
        \begin{algorithmic}[1]
        \Statex [bool, CPR]$ = \textit{\changedSB{cut}PointCheck}~(w,v', G_N)$
    \Statex \hrulefill
         \Statex Inputs:
         \begin{enumerate}[leftmargin=2em,label=\alph*.]
         \item Two vertices with intersecting neighborhoods, $w \in V_N, v'$ ($v'$ is not inserted in the graph)
         \item Parent vertex of $v'$, $v \in V_N$
         \item Neighborhood augmented graph, $G_N$
        \end{enumerate}
    \Statex Output: 
         \begin{enumerate}[leftmargin=2em,label=\alph*.]
            \item  Boolean (\textbf{true} if \changedSB{cut} point, \textbf{false} if not)
            \item  \changedSB{Cut} point region (a set of vertices), CPR
        \end{enumerate}
    \Statex \hrulefill
    \State $\alpha = $ \textit{getNeighborhoodIntersectionRatio}~$(w.\mathcal{U},v'.\mathcal{U})$ \label{ln:getNeighborhoodIntersectionRatio}
    \If {$\alpha > \varepsilon_{i}$} 
        \State return (\textbf{false},$\emptyset$)
    \EndIf
    \If {$|g(w.\text{came\_from}) - g(v)| > \varepsilon_g$} \label{ln:gscorecheck}
        \State return (\textbf{false},$\emptyset$)
    \EndIf
    \State $p_w = $ \textit{getPathPoint}~$(w,r_l, G_N)$\algcomment{\textit{getPathPoint}~$(w,r_l, G_N)$ returns the vertex at a distance of $r_l\cdot g(w)$ from $w$, along the shortest path leading up to the start vertex} \label{ln:ppw}
    \State $p_{v'} = $ \textit{getPathPoint}~$(v',r_l, G_N)$ \label{ln:ppv'}
    \State $d_{sep} = $ \textit{searchOnGraph}~$(p_w,p_{v'},G_N)$ \algcomment{by running A* from $p_w$ to $p_{v'}$ with a radius of $\varepsilon_{upper}$, returns the distance if $p_{v'}$ is reached, returns \textbf{false} otherwise} \label{ln:dsep}
    \If {$d_{sep}$ \textbf{AND} $d_{sep} < \varepsilon_{upper} $}\label{ln:boundcheckMP}
        \State MPR $= $ \textit{generateCutPointRegion}~$(w,r_{MP}, G_N)$ \algcomment{by running Dijkstra's starting from $w$ with a radius $r_{MP}$, expanded vertices are included within \changedSB{cut} point region}
        \State return (\textbf{true},MPR)
    \EndIf
\end{algorithmic} 
\end{algorithm}

%% file: sections/length_constrained_search.tex
\subsection{Problem Description}
Tethered robots are mostly deployed to perform tasks in environments where wireless communication is limited and/or robot is unable to operate without an external power source. This work is mainly interested in applications where a tethered spatial robot is to navigate in a 3D environment with obstacles. Examples include a drone tethered to an outside base navigating in and around a building with windows or gates or a disaster site with multiple entries and exits, an underwater robot tethered to the surface station navigating in and around a ship wreck or an underwater cave with tunnels and passages. In the context of path planning, the aim is to find a path between a start and a goal configuration in the configuration space of the tethered robot, that will satisfy the physical constraints brought by the tether.

\begin{figure}
    \centering
    \includegraphics[width=0.5\linewidth]{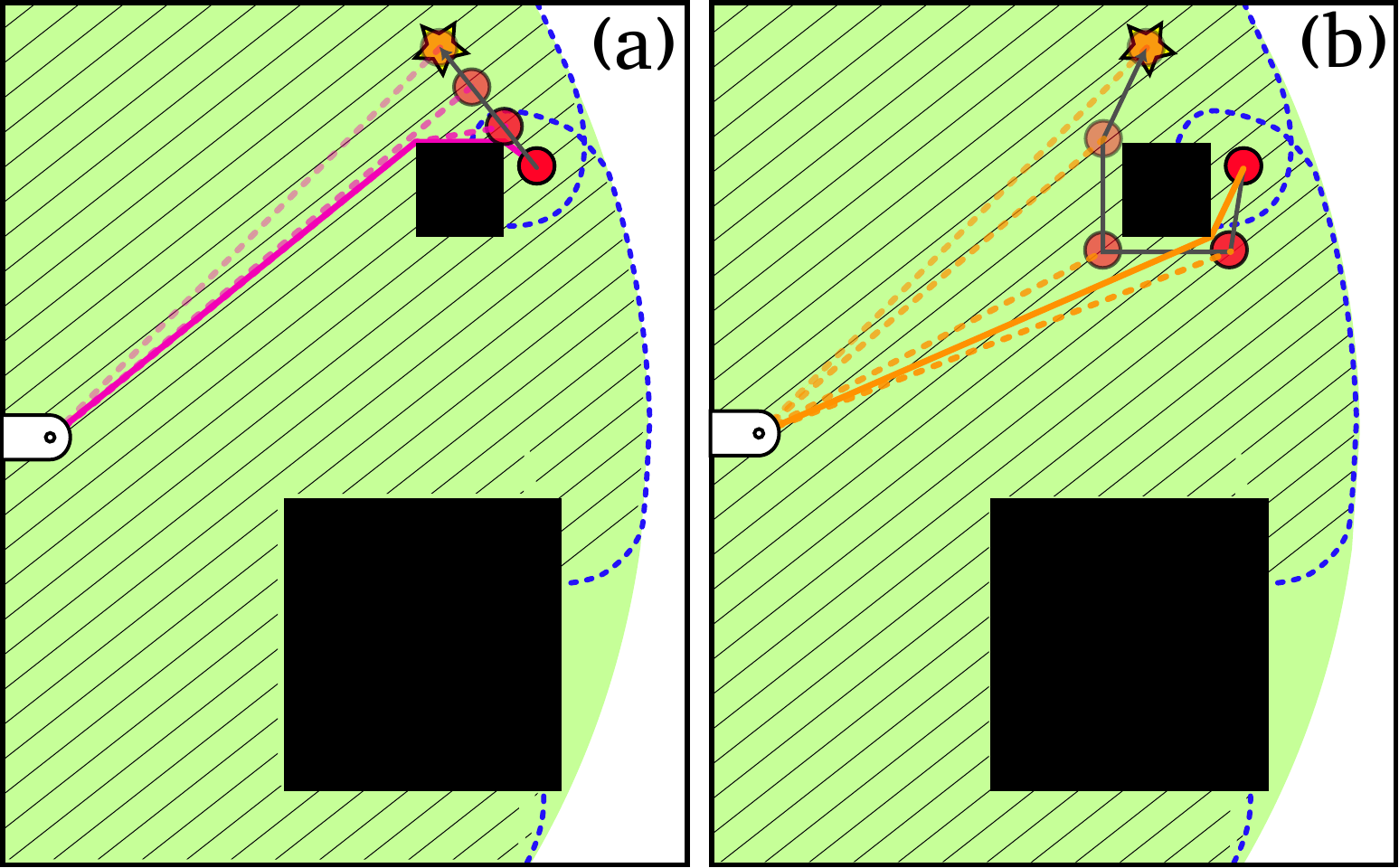}
    \caption{(a) Tethered robot starts from an initial configuration and reaches the goal via the \textit{globally shortest} length constrained path. (b) Robot starts from at the same coordinates but with a different tether configuration. It reaches the same goal via another length constrained path that is \textit{locally shortest}.}
    \label{fig:lcs_description}
\end{figure}

For the purposes of this paper, it can be assumed that the tether is made from an elastic material with a maximum length $l$, which remains taut during the entire operation irrespective of the location of the robot. Alternatively, it can be assumed that there is a tether retraction mechanism at the base, that can remove the remaining slack from the tether when it is not used at its maximum length $l$. Using any of the two assumptions, a tether configuration can be expressed as one of the locally shortest paths from the tether base to robot's location. An illustration of a tethered robot in a 2D environment with obstacles is given in Figure~\ref{fig:lcs_description}.

As observed in Figure~\ref{fig:lcs_description}, being tethered to a fixed base constrains the workspace of the robot, which could otherwise contain the entire environment. The region colored in green contains the points that are within a radius of $l$ to the base, where the white colored region lies out of the workspace of the robot. Obstacles in the environment, constrain the workspace even further for a tethered robot (from the green colored region to the shaded one surrounded by the blue dashed lines), as the tether is not expected to go through the obstacles and it has to be stretched longer around the obstacle. As seen around the smaller obstacle, there also exist points in the environment which can be reached via different tether configurations within the length constraint. 


\subsection{Solution} \label{subsec:length_constrained_search}

\changedSB{In prior work we developed a method for computing shortest traversible path in a planar, Euclidean domain with obstacles for a tethered robot with cable length constraint using a topological path planning (homotopy-augmented graph) method~\cite{Kim2014}. This approach, however, cannot be naturally extended to 3D domains since in spatial domains with obstacle there may exist locally-optimal paths that belong to the same homotopy class, but need to be made distinction between because a cable with length constraint cannot be continuously deformed from one topo-geometrically distinct path to another. Even for different topological classes, computation of homotopy invariants is extremely difficult in spatial domains~\cite{Homotopy-Planning:journal:18}. Equipped with neighborhood-augmented graph for topo-geometric path planning, we can now compute optimal traversible path for a tethered robot with cable length constrain in complex spatial domains.}

Path planning for a tethered robot consists of two stages of search using neighborhood augmented graphs. The first stage is referred to as \textit{workspace exploration}, where a neighborhood augmented graph is incrementally constructed starting from a base vertex $v_b$ to each point in the environment that is reachable within the tether length constraint. The algorithm used for \textit{workspace exploration} is identical to the one in Algorithm~\ref{alg:NASearch}, where the stopping criteria is a radius based one. Instead of stopping at a vertex that has matching coordinates with the assigned goal, \textit{workspace exploration} terminates at the first vertex that has a g-score exceeding the maximum cable length (the stopping criterion is $g(v) < l$, similar to the one in the \textit{computeTNS} routine given in Algorithm~\ref{alg:computeNeighborhood}). During the construction of this fixed length graph, we use the S* algorithm, such that the computed g-scores represent the lengths of the shortest paths in the underlying configuration space. 

The second stage is referred to as \textit{length constrained search (LCS)}, where another search is performed on the previously constructed neighborhood augmented graph. This search starts from an initial vertex $v_s$ (note that on a neighborhood augmented graph there could exist multiple vertices with identical coordinates, but with distinct neighborhoods), which also corresponds to a unique tether configuration in the neighborhood augmented graph, and terminates at a goal coordinate $q_g$. In this case, there are no topological or geometrical constraints on the final tether configuration placed within the search algorithm, thus any vertex $v$ that has matching coordinates with $q_g$ is a valid goal vertex and any corresponding tether configuration is valid. However, as observed in Figure~\ref{fig:lcs_description}, 
the shape of the workspace dictates itself during the \textit{LCS} as it is performed on the graph of the explored workspace, thus some of the goal coordinates might only be reached via certain tether configurations. Path resulting from a \textit{LCS} is referred to as \textit{length constrained path (LCP)} and is a locally shortest path between the start and the goal coordinates that satisfies the tether length constraint. It must be emphasized that the \textit{LCP} obtained using this method is the globally optimal path among all the alternative paths that satisfy the tether length constraint, but it may be only locally optimal in the unconstrained configuration space.


\subsection{Tether Length Analyis}
\begin{figure}[h!]
    \centering
    \includegraphics[width=\linewidth]{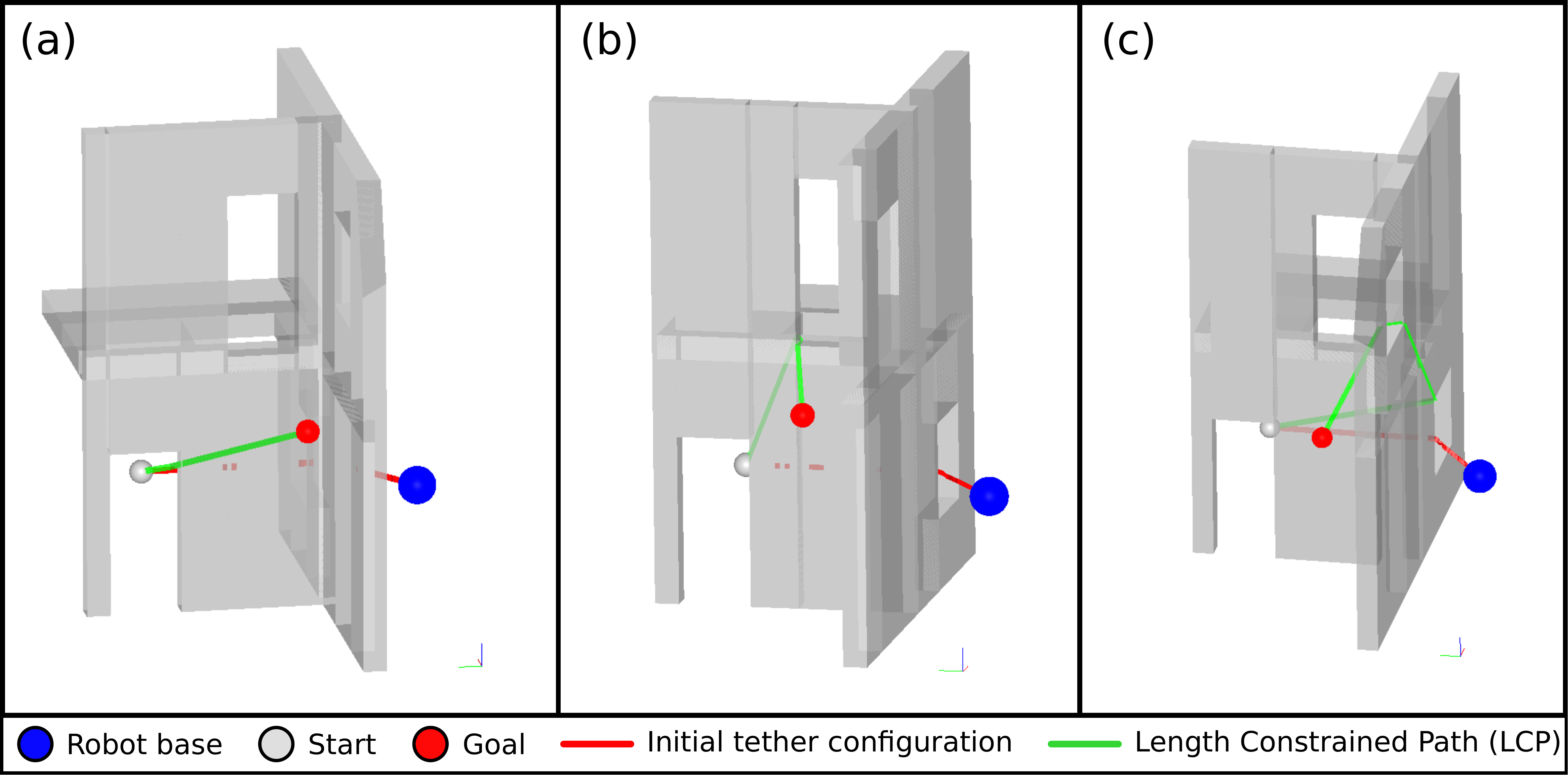}
    \caption{\textbf{Results:} (a) Max. tether length $= 30$. Globally shortest path in the environment satisfies the tether length constraint. (b) Max. tether length $= 25$. With a shorter tether, robot has to follow the next locally shortest path (which is the second globally shortest path in the environment). (c) Max. tether length $= 20$. With a stricter constraint on the tether length, the path that the robot has to follow is significantly longer.}
    \label{fig:cable_length_analysis}
\end{figure}

To demonstrate the effect of the tether length constraint on the paths obtained by the length constrained search algorithm, we consider a building-like environment shown in Figure~\ref{fig:cable_length_analysis}. The building has a total of 3 rooms, where the first room on the left spans two floors and remaining two rooms are on the right connected via a hole in between. Both rooms on the right connect to the first room via doors. The first room and the room at bottom right are accessible from outside via windows. Robot is tethered to a base fixed outside and initially positioned inside the bottom right room, which it accessed through the corresponding window. The goal is placed within the first room near the window.

Keeping the initial and goal configurations the same, the length constrained search algorithm is run with varying maximum tether length $l$. As observed in Figure~\ref{fig:cable_length_analysis}, \textit{LCS} algorithm outputs the shortest path between the start and goal that would satisfy the tether length constraint. For relatively optimistic/relaxed constraints, resulting path might correspond to the globally shortest path in a given environment. With stricter constraints, the algorithm is still able to find a path between the start and goal. However, these paths might be significantly longer, requiring the robot to exit the building through the windows that it has already used while entering.

\subsection{Simulations} \label{sec:length-constrained-search-simulation}
\begin{figure*}[h!]
    \centering
    \includegraphics[width=\linewidth]{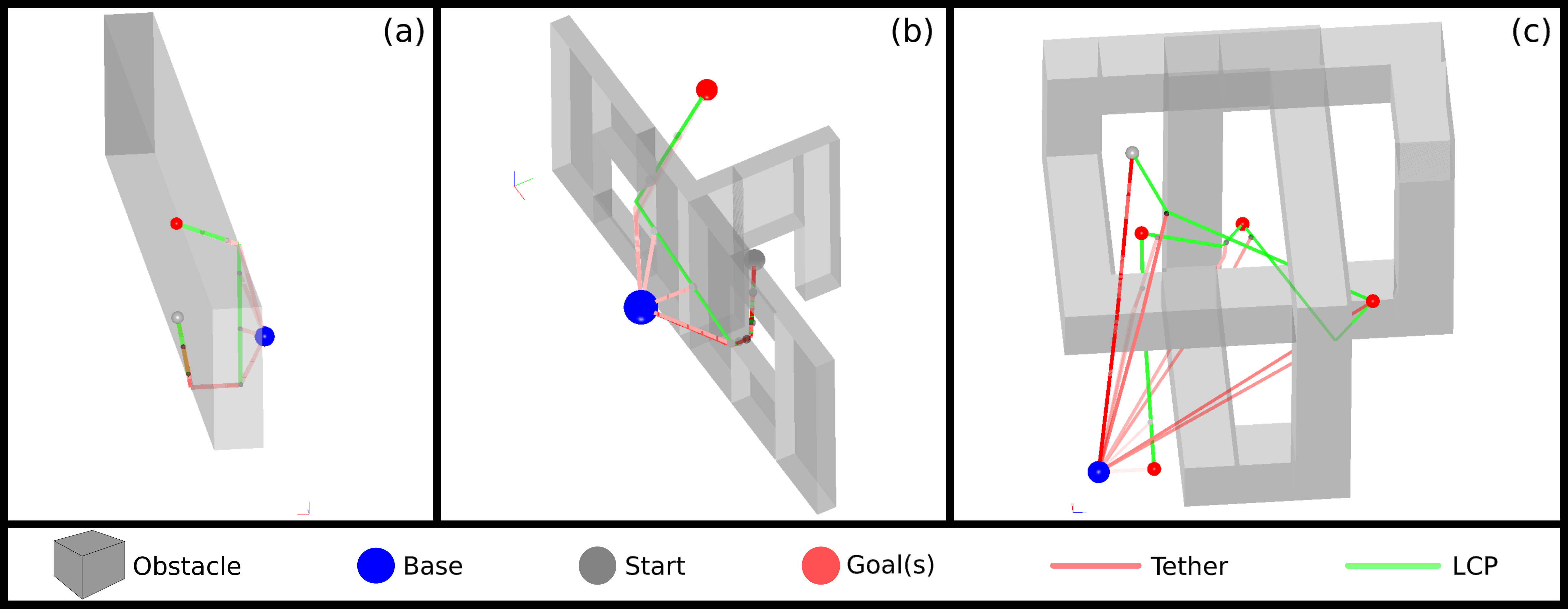}
    \caption{\textbf{Results:} (a) Environment with long rectangular obstacle. (b) Environment with building-like structure. Building has two rooms connected via a door. Both rooms have window access and the robot base is placed outside. (c) Environment with a trefoil knot shaped obstacle. Sequence of tether configurations are plotted starting from opaque to transparent red to highlight the order.}
    \label{fig:lcs_simulations}
\end{figure*}

We demonstrate the capabilities of the proposed \textit{LCS} algorithm on virtual environments shown in Figure~\ref{fig:lcs_simulations} \changedASnew{and the accompanying multimedia attachment}. In the scenario shown in Figure~\ref{fig:lcs_simulations}(a), there exists a long rectangular prism in the environment around which paths would be topologically equivalent but geometrically distinct. The robot is to start from an initial position and a corresponding tether configuration such that the globally shortest path from the start to goal does not satisfy the tether length constraint. The \textit{LCP} obtained from the algorithm is shown using the green line and the sequence of tether configurations are shown using the transparent red lines. As expected, the \textit{LCP} makes the robot move towards its base and go around the other side of the prism. It must be noted that in this scenario the path planning takes place around a genus-0 prism, which was described amongst limiting cases in the previous sections. The difference is that beyond a certain ratio of the length, width and depth of a prism, it is possible to rely on the original version of the algorithm (proposed in Sections~\ref{subsec:neighborhood_augmented_graph},\ref{subsec:neighborhood_generation}) to make a geometrical distinction across the faces instead of the corners of the prism.

A similar behavior is observed in Figure~\ref{fig:lcs_simulations}(b), where the robot has to exit the building-like structure and use the alternative entrance instead of taking the globally shortest path to reach the goal. Figure~\ref{fig:lcs_simulations}(c) features a trefoil knot shaped obstacle around which a robot is to navigate between a series of goal points. In this scenario, the robot follows the globally shortest paths whenever they satisfy the tether length constraint and uses the alternative locally shortest paths when necessary.

\subsection{Real Robot Experiments} \label{sec:length-constrained-search-experiment}
\begin{figure*}[h!]
    \centering
    \includegraphics[width=\linewidth]{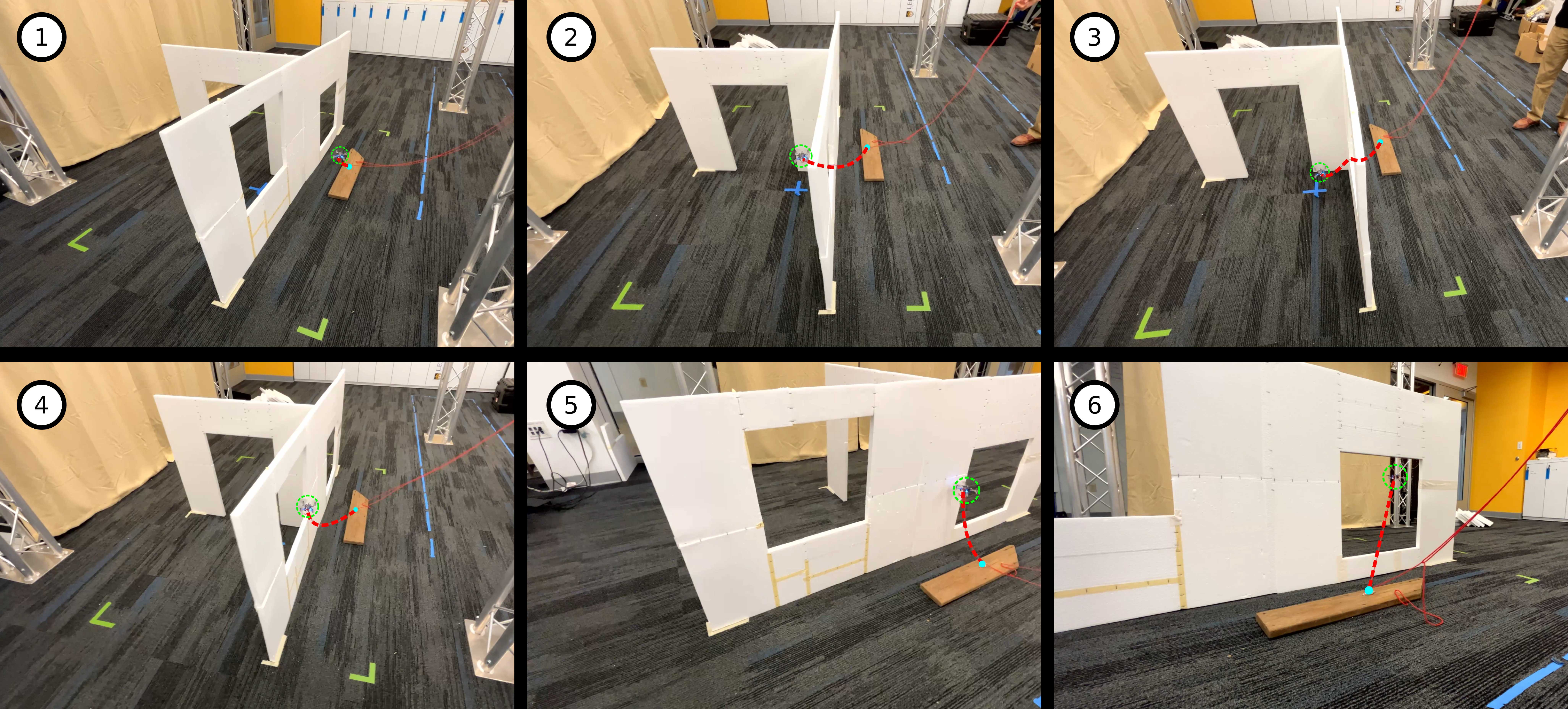}
    \caption{Photos from real robot experiments for scenario 1a. Steps 1 through 6. Quadrotor, wool tether and robot base are highlighted with colored lines for improved visibility.}
    \label{fig:lcs_experiment_steps}
\end{figure*}

\changedASnew{A structure similar to the one shown in Figure~\ref{fig:lcs_simulations}(b) is constructed using styrofoam blocks for real robot demonstrations. A Crazyflie 2.1 quadrotor platform \cite{bitcrazeCrazyflieBitcraze} is connected to a metal hoop (that acts as a robot base) via a red wool thread (that represents the tether). Discrete output obtained from the path planning algorithm is used to generate a feasible trajectory for the quadrotor using the implementation in \cite{whoenig/uav_trajectories} based on \cite{richter2016polynomial,burri2015real-time}. Resulting trajectories are tracked using the controllers proposed in \cite{mellinger2011} along with an OptiTrack motion capture system within the Crazyswarm framework \cite{crazyswarm}.

The experimental setup and the navigation progress is to be seen in Figure~\ref{fig:lcs_experiment_steps} and the accompanying multimedia attachment. The quadrotor is initially placed on the ground closer to the window on the left. The first goal is defined inside the room on the left which is accessed via the window as shown in Figure~\ref{fig:lcs_experiment_steps} - Step 2 and 3. The second goal is defined inside the room on the right. The globally shortest path to the second goal is not traversable due to the tether length constraint.  As shown in Steps 4-6, the quadrotor has to exit the structure and use the window on the right to reach the goal via a locally shortest path without violating the constraint.}

%% file: sections/discussion_conclusion.tex
We present a topo-geometric path-planning approach for finding multiple, locally optimal paths. This is accomplished by constructing neighborhood-augmented graphs in which every vertex is augmented with a neighborhood around the path leading up to that vertex. Existing graph-search algorithms can be run on these graphs to generate desired number of paths. Our method is general in the sense that it provides a way to compute multiple locally-optimal paths in configuration spaces of various geometry and dimension without requiring any additional constructions or procedures. Unlike existing topological path planning methods, it is also able to find locally optimal paths that are homotopic but geometrically different.

Path planning capabilities of the algorithm are demonstrated on a variety of environments and scenarios. We further discussed the limitations of the proposed approach in low-curvature environments and present modifications to the algorithm to increase path planning capabilities in these type of environments. Lastly, we demonstrated the use of the neighborhood-augmented planning algorithm for a length constrained path planning problem. For a tethered robot navigating in 3D, we find optimal paths that satisfy the tether length constraints, which we demonstrate using simulations and real robot experiments.